\documentclass{article}

\usepackage[preprint]{nips_2018}

\usepackage[utf8]{inputenc} 
\usepackage[T1]{fontenc}    
\usepackage{hyperref}       
\usepackage{url}            
\usepackage{booktabs}       
\usepackage{amsfonts}       
\usepackage{nicefrac}       
\usepackage{microtype}      
\usepackage{graphicx}
\usepackage{amsmath}
\usepackage{multirow}
\usepackage{algpseudocode}
\usepackage{makecell}
\usepackage{subfigure}
\usepackage[export]{adjustbox}

\title{Concurrent Meta Reinforcement Learning}

\author{%
  Emilio Parisotto \\ {\small Machine Learning Department} \\ {\small Carnegie Mellon University} \\ {\small Pittsburgh, PA 15213} \\ \texttt{\small eparisot@cs.cmu.edu} \And
  Soham Ghosh \\ {\small Machine Learning Department} \\ {\small Carnegie Mellon University} \\ {\small Pittsburgh, PA 15213} \\ \texttt{\small sohamg@cs.cmu.edu} \And
  Sai Bhargav Yalamanchi \\ {\small ECE Department} \\ {\small Carnegie Mellon University} \\ {\small Pittsburgh, PA 15213} \\ \texttt{\small syalaman@andrew.cmu.edu} \AND
  Varsha Chinnaobireddy \\ {\small ECE Department} \\ {\small Carnegie Mellon University} \\ {\small Pittsburgh, PA 15213} \\ \texttt{\small vchinnao@andrew.cmu.edu} \And
  Yuhuai Wu \\ {\small Department of Computer Science} \\ {\small University of Toronto and} \\ {\small Vector Institute} \\ \texttt{\small ywu@cs.toronto.edu} \And
  Ruslan Salakhutdinov \\ {\small Machine Learning Department} \\ {\small Carnegie Mellon University} \\ {\small Pittsburgh, PA 15213} \\ \texttt{\small rsalakhu@cs.cmu.edu} \\
}

\begin{document}

\maketitle

\begin{abstract} 
  State-of-the-art meta reinforcement learning algorithms typically assume the setting of a single agent interacting with its environment in a sequential manner. A negative side-effect of this sequential execution paradigm is that, as the environment becomes more and more challenging, and thus requiring more interaction episodes for the meta-learner, it needs the agent to reason over longer and longer time-scales. To combat the difficulty of long time-scale credit assignment, we propose an alternative parallel framework, which we name ``Concurrent Meta-Reinforcement Learning'' (CMRL), that transforms the temporal credit assignment problem into a multi-agent reinforcement learning one. In this multi-agent setting, a set of parallel agents are executed in the same environment and each of these "rollout" agents are given the means to communicate with each other. The goal of the communication is to coordinate, in a collaborative manner, the most efficient exploration of the shared task the agents are currently assigned. This coordination therefore represents the meta-learning aspect of the framework, as each agent can be assigned or assign itself a particular section of the current task's state space. This framework is in contrast to standard RL methods that assume that each parallel rollout occurs independently, which can potentially waste computation if many of the rollouts end up sampling the same part of the state space. Furthermore, the parallel setting enables us to define several reward sharing functions and auxiliary losses that are non-trivial to apply in the sequential setting. We demonstrate the effectiveness of our proposed CMRL at improving over sequential methods in a variety of challenging tasks.
\end{abstract}

\section{Introduction}

\begin{figure}[ht!]
    \centering
    \includegraphics[width=1\linewidth]{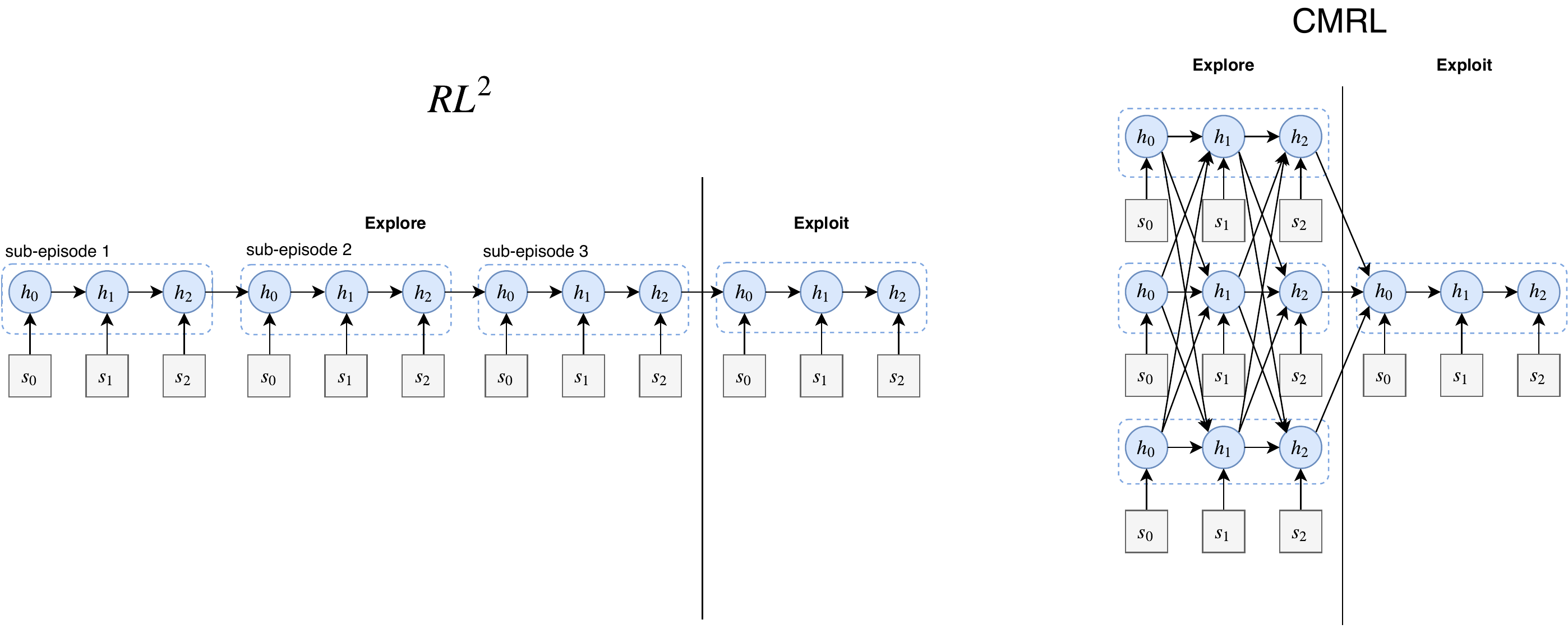}
    \caption{Our proposed Concurrent Meta Reinforcement Learning (CMRL) framework (right) and the sequential setting common to most previous meta reinforcement learning algorithms, such as RL2 (left).}
    \label{fig:my_label}
\end{figure}

Current deep reinforcement learning algorithms have achieved a wide variety of successes~\citep{silver2017masteringgo,moravvcik2017deepstack} but suffer from poor sample complexity. 
The extension of Deep Reinforcement Learning's success to the few-shot setting could enable its utilization in a wide range of previously inaccessible applications, where either accurate simulators are not available or large amounts of data are intractable to acquire.
Owing to the large practical significance of any potential few-shot reinforcement learning algorithm, a large amount of recent work has focused on this topic. 
One recent research direction within this broader field concerning improving deep reinforcement learning's sample efficiency is to examine whether it is possible to learn the learning algorithm itself.
Learning the learning algorithm would loosen the constraint that the method should be successful on a general space of problems (e.g. all Markov Decision Processes), and it could instead focus on adjusting its biases so that it can adapt as quickly as possible on more targeted structured domains, e.g., only environments with dynamics resembling real-world physics.
The promise of meta-learning algorithms is therefore, by this reduction in generality and specialization to a set of tasks, to create an agent that can produce optimal behaviour within a few-shot number of interaction episodes within its environment (i.e. on the order of 10). 

The methods which have emerged from the recent work on meta learning in the reinforcement learning setting can be classified into two broad categories, meta-optimization and episodic memory.
Meta-optimization approaches meta-learning in the context of optimization algorithms, where certain properties of the model are attained by backpropagating through the optimization algorithm itself. These meta-optimization methods have demonstrated impressive performance in the areas of reinforcement learning~\cite{finn2017model}, learning to explore~\cite{gupta2018meta}, unsupervised reinforcement learning~\cite{gupta2018unsupervised} and multiagent and non-stationary environments ~\cite{al2017continuous, raileanu2018modeling}. 
The second main meta reinforcement learning approach is through the use of an inter-episodic memory. In contrast to typical memory architectures which refresh memory states at episode boundaries, the use of a persistent inter-episodic memory functions to allow the agent to not only reason about what is happening in its current episode, but what it might have tried in previous interactions with its environment. In contrast to meta-optimization methods where the inter-episodic memory is essentially the adapting weights of the model, episodic meta reinforcement learning achieves rapid adaptation of behaviour by changes to its memory state.

Both of the aforementioned approaches to Meta Reinforcement Learning have a commonality whereby the few-shot meta-learning agent is given a set of interaction episodes in sequence, and it conditions itself (whether by adapting its weights or changing its memory state) on its past experience to produce new behaviour on the next interaction. The sequential setting has a potential difficulty in scaling to challenging domains where a large number of interactions are required to solve the environment. One reason for this is that the method of temporal credit assignment within either of the main meta reinforcement learning approaches, backpropagation through time, is known to degrade as sequence lengths grow longer~\cite{bengio1994learning} and often requires specialized architectures to enable effective optimization. This temporal credit assignment problem is further amplified in the reinforcement learning setting as individual episodes can be on the order of several hundred time steps long. Furthermore, the sequential setting is not always desirable or even feasible, such as in applications where time is a considerable expense. For example, in a real-world robotics application, a meta-learner in the sequential setting could use at most one robotic arm at a time. Given that each single episode within a few-shot interaction might take a significant amount of time, a meta-learner that can aggregate multiple streams of information in an online fashion would be more efficient as it is capable of using multiple arms at once.

To address these challenges, we propose ``Concurrent Meta Reinforcement Learning''  (CMRL),  an alternative framework to the sequential setting commonly assumed in meta reinforcement learning. Here, an agent does not necessarily interact with an environment in a strict sequence of interaction episodes, but runs several episodes in parallel and then aggregates this information to produce an optimal policy.
This alternative parallel mode of interaction is a general framework that can be adapted to many of the  meta reinforcement learning algorithms previously devised in the sequential setting. In this paper, we focus on the episodic reinforcement learning meta-learners, particularly those methods outlined in RL2~\cite{duan2016rl} / L2RL~\cite{wang2016learning} and the later extension of ERL2~\cite{stadie2018some}. In the concurrent episodic meta reinforcement learning setting, a set of "rollout" agents are executed from the same initial state but each agent is given the means of communicating  with the others, enabling coordination between the agents on which regions of the state space to explore. A key benefit to the use of communicating parallel rollouts instead of sequential episodes is the enabling of the use of various reward sharing schemes, which can both encourage the rollouts towards exploitative behaviour but also not discourage higher risk exploratory actions. Additionally, several divergence losses between the parallel policies can be easily defined allowing a further push towards varied behaviours. We demonstrate the effectiveness of our concurrent meta reinforcement learning framework in a simple graph-based MDP, a partially-observable maze navigation task and a robotic arm manipulation task.

\section{Background}

We denote by $\mathcal{T} = (\mathcal{S}, \mathcal{A}, \mathcal{P}, \gamma, \mathcal{R}, \rho, T)$ a Markov Decision Process, where $\mathcal{S}$ is the finite set of states, $\mathcal{A}$ is the finite set of actions, $\mathcal{P} : \mathcal{S} \times \mathcal{A} \times \mathcal{S} \rightarrow [0,1]$ is the transition kernel, $\gamma$ is the discount factor, $\mathcal{R} : \mathcal{S} \rightarrow \mathbb{R}$ is the reward function, $\rho : \mathcal{S} \rightarrow [0,1]$ is the initial state distribution and $T$ is the time horizon after which an episode is guaranteed to end. A stochastic policy $\pi : \mathcal{S} \times \mathcal{A} \rightarrow [0,1]$ maps state-action pairs to probabilities. The value function $V^\pi$ of a policy $\pi$ represents the discounted future reward if we started from a given state and continued executing the policy: $V^\pi(s) = \mathbb{E}_{\pi,\mathcal{P}}\left[\sum_{t=0}^T \gamma^t r_t\right]$.  The goal of reinforcement learning is to produce a paramaterized policy $\pi_\theta$ that maximizes the discounted future reward over states sampled from the initial state distribution:
\begin{align*}
    L_\mathcal{T} = \mathbb{E}\left[\sum_{t=0}^T \gamma^t r_t \middle| s_0\sim\rho,a_t\sim\pi(s_t),s_{t+1}\sim\mathcal{P}(s_t,a_t) \right]
\end{align*}
We use the Advantage Actor Critic (A2C)~\cite{mnih2016asynchronous} algorithm to optimize over this objective, parameterizing the policy using a deep network. 

\subsection{Meta Reinforcement Learning}

Meta Reinforcement Learning is concerned with the case where we are given a distribution $p : \mathcal{T} \rightarrow [0,1]$ over tasks $\mathcal{T} = \{\mathcal{T}_1,\ldots,\mathcal{T}_N\}$ and we want to learn how to act in any environment sampled from this distribution as quickly as possible. By learning to focus on only the variations between the environments in $p(\mathcal{T})$, meta learning algorithms can potentially exploit structure in the class of environments $\mathcal{T}$ to, in the most extreme case, learn a new  $\mathcal{T}_i\sim p$ in only 1-10 interaction episodes. 

Adopting similar conventions to~\cite{finn2017model,stadie2018some}, we can frame meta reinforcement learning as a solution to the following objective function:
\begin{align*}
    \min_{\theta} \sum_{\mathcal{T}_i} \mathbb{E}_{\pi_{\Delta(\theta)}}\left[\mathcal{L}_{\mathcal{T}_i}\right]
\end{align*}
Where $\Delta(\theta)$ represents a few-shot update method which collects a limited amount of experience from each $\mathcal{T}_i$ to update $\theta$. 

\subsection{Meta-Episodic Structure}
\label{sec:exploreexploit}

In this section, we standardize the episodic meta-learning algorithms within a common framework. We call a ``meta-episode'' the set of interactions a meta-learner has with a particular instance of an environment class $\mathcal{T}$. Each interaction episode within this meta-episode is called a sub-episode. We assume that there are $K$ sub-episodes in a meta-episode, and we further sub-divide these sub-episodes into two types of categories. The first type is the explore or exploratory sub-episodes, of which there are $K_{explore}$. The second type is the exploit or exploitative sub-episodes, which there are $K_{exploit}$. We have that $K = K_{explore} + K_{exploit}$. In this work, we only consider settings in which there is only 1 exploit sub-episode.

Furthermore, in contrast to previous work on inter-episodic memory, we have separate RNN weights and policy outputs between explore and exploit sub-episodes. This makes RL2 more closely resemble a sequence-to-sequence model~\cite{sutskever2014sequence}, as it first encodes exploratory sub-episodes into a fixed-vector and then uses that code vector as the initial state of an exploitative recurrent policy. While a minor change, the separation of explore and exploit policy parameters can enable an easier optimization for the meta-learner to learn to produce potentially radically different behaviour between explore and exploit sub-episodes. 

\section{Related Work}

Meta reinforcement learning is a rapidly developing field, with a large amount of recent work addressing the problem due to the potentially far-reaching practical significance of reducing deep reinforcement learning's sample complexity. The meta reinforcement learning methods can be broadly classified into two separate sub-groups: optimization-based methods and episodic-memory-based methods. 

The optimization-based meta learning methods, or meta-optimization methods, typically incorporate some form of SGD as a component of the meta-learner's architecture, backpropagating through the optimization procedure itself to optimize some initial parameters. For example, MAML~\cite{finn2017model} optimizes an initial policy weight matrix so that it is ``close'' to the optimal policy weights for any MDP sampled from a class of tasks, where closeness denotes that it only takes a few SGD updates to reach the optimum starting from the trained initial weights. MAML was shown to be more successful than traditional fine-tuning methods on a variety of supervised and 2D navigation tasks. More recent work has extended this meta-optimization idea into more specialized facets of reinforcement learning, such as non-stationary and multiagent domains~\cite{al2017continuous}, learning to explore~\cite{gupta2018meta}, and recovering agent's latent goals~\cite{raileanu2018modeling}. 

An alternative approach to meta-optimization methods are those focusing on episodic memory, or memory that persists between episodes. These algorithms utilize memory architectures such as LSTMs as the means to transmit and retain information over episodes, This allows the agent to remember behaviours it has tried and reason over its own successes and failures to produce more optimal behavior on a later interaction.
Early work on this approach used LSTM architectures as the inter-episodic memory~\cite{wang2016learning,duan2016rl} and, despite the simplicity of its architecture, demonstrated success on a variety of bandit and 3D visual navigation tasks. While effective, RL2 is not capable of solving certain classes of MDPs optimally if exploration has a significant risk of negative reward (see Figure~\ref{fig:unsolvable} and surrounding text for a counter-example). Later work~\cite{stadie2018some} extended the RL2 framework to be able to handle these higher risk environments by introducing a notion of ``explore''  and ``exploit'' sub-episodes within the meta-episode, allowing only external reward signals within exploit sub-episodes to affect the policy optimization. This extension, termed ERL2 for Exploratory-RL2, enabled the policy to take more risky behaviour during exploration at the cost of a more difficult temporal credit assignment problem for the meta-learner as external signals were further sparsified. 

A parallel line of work has also looked at more non-parametric forms of episodic memory, where percepts are stored directly in a database and an agent does retrieval on the information stored in order to decide how to act. These database-based approaches more closely resemble traditional forms of tabular reinforcement learning, but typically utilize a non-parametric kNN-style indexing mechanism due to the high-dimenisonal state space of many important applications. The Episodic Control~\cite{blundell2016ec} algorithm used a memory bank of stored states, actions and values acquired as the agent acts within the environment, and from which new actions are chosen using a nearest-neighbor look-up of the stored Q-values of previous states. Episodic Control's nearest-neighbour look-up used features that were either random projections or pre-trained using variational autoencoder losses. Later work extended the Episodic Control framework to settings where features~\cite{pritzel2017nec} and stored values~\cite{ritter2018been} are jointly learned with the agent's policy rather than being hand-engineered.
    
A related work in the non-meta-learning setting is that of Concurrent Reinforcement Learning~\cite{pmlr-v28-silver13}. This work proposed a new concurrent setting beyond the typical sequential or continual settings used in most reinforcement learning formalizations. The concurrent setting is defined as the case where episodic interactions occur in parallel, and~\cite{pmlr-v28-silver13} designed and analyzed a temporal difference update to handle learning in this setting. Concurrent Reinforcement Learning has many similarities to our proposed Concurrent Meta Reinforcement Learning in the sense that information from multiple parallel rollouts must be aggregated into a cohesive optimal behaviour, but Concurrent Reinforcement Learning pre-defines the update rule whereas CMRL aims to learn them. Concurrent Reinforcement Learning also dealt with the case of asynchronous actions, which CMRL currently does not explicitly handle. We leave this extension to future work. More recent work in the concurrent setting has looked at applying techniques from the Posterior Sampling for Reinforcement Learning (PSRL) literature~\cite{strens2000bayesian} to the concurrent setting~\cite{dimakopoulou2018coordinated,dimakopoulou2018scalable}, with recent extensions using neural network function approximators in a sparse cartpole domain~\cite{dimakopoulou2018scalable}. These approaches are fundamentally different from ours in that concurrent learning is not dealt with as a deep multi-agent RL communication problem but as a method of efficiently sharing information about models and coordinating exploration based on the shared posterior of the model.

Recently a wide variety of works have examined the question of multi-agent communication in the Deep Reinforcement Learning setting. Early work in communication architectures used relatively simple mechanisms: CommNet~\cite{sukhbaatar2016learning} used an iterated average pooling of other agents features to transmit information between agents, whereas ~\cite{foerster2016learning} had a dedicated communication channel between two agents that was either discrete or continuous, with potential channel noise added. These early architectures showed success on a variety of simple environments such as a traffic junction simulation. VAIN~\cite{hoshen2017vain} extended CommNet with an attention mechanism that allowed agents to focus on the features of particular agents it wished to receive information from. 

\section{Concurrent Meta Reinforcement Learning}
\label{sec:cmrl}

\begin{figure}
  \centering
  \begin{minipage}{0.35\linewidth}
    \includegraphics[width=1.00\linewidth]{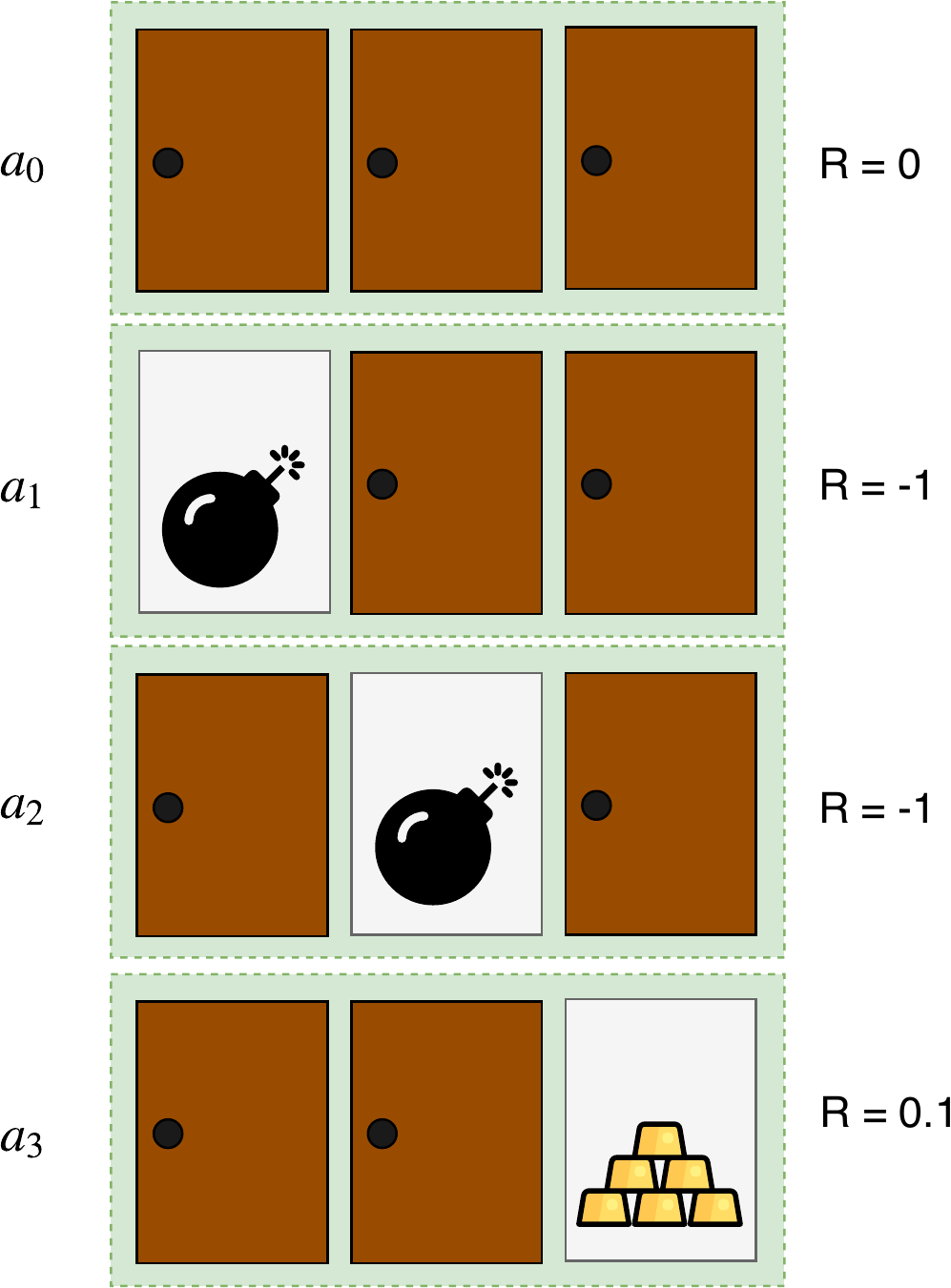}
  \end{minipage}%
  \begin{minipage}{0.05\linewidth}
    $\quad$
  \end{minipage}%
  \begin{minipage}{0.35\linewidth}
    \includegraphics[width=1.00\linewidth]{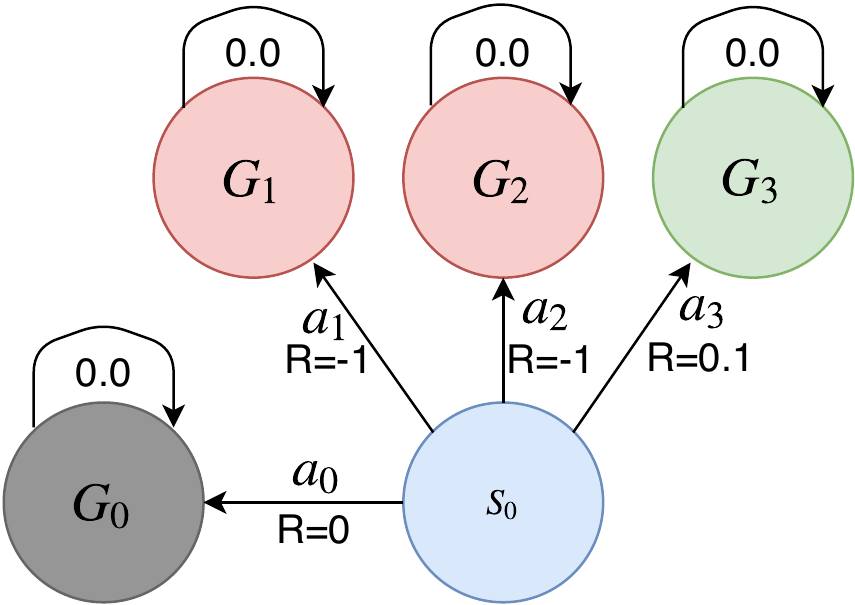}
  \end{minipage}
  \caption{An example illustration (left) and its MDP (right) of a particular instance environment of the Monty-Hall class of MDPs. From the starting state $s_0$, the agent is given a choice of 4 actions. The first action is a NOOP and the action ends the episode without reward. The other actions open one of the 3 doors, where 2 of the doors lead to large negative rewards (illustrated with a cartoon bomb) and one door leads to a positive reward (represented by the gold bars). This class of MDPs is unsolvable using naive RL2 because it must in the worst case incur more negative reward during exploration sub-episodes than it can recover in the exploitation sub-episode.}
    \label{fig:unsolvable}
\end{figure}

Meta Reinforcement Learning typically processes episodes in a sequential manner, conditioning on all past history to influence the behaviour within the next interaction episode. While the sequential processing of episodes has been shown to be effective~\cite{duan2016rl,wang2016learning,stadie2018some}, it has several drawbacks. One of these drawbacks is that the standard mode of learning within these algorithms, backpropagation through time, degrades in performance significantly as the time scales increase due to an issue referred to as vanishing/exploding gradients~\cite{bengio1994learning}. Depending on the meta-time horizon of the environment, which is a product between the number of interaction sub-episodes available and the time horizon of each individual sub-episode, this can severely limit the scalability of sequential-based meta reinforcement learning methods to more challenging environments where a few-shot number of interactions might not be sufficient for learning optimal behaviour.
 
Furthermore, there are a number of limitations to the types of environments that RL2 and related episodic meta-reinforcement learning algorithms can solve. For a concrete counter-example, consider a simple class of MDPs with 1 initial state and 4 absorbing states, shown in Figure~\ref{fig:unsolvable}. The initial state can transition to any absorbing state. Transitioning to three of the four absorbing states represent goals providing a reward, and the other absorbing state represents a NOOP that provides no reward. Suppose that within this class of MDPs (which we call the Monty-Hall class of environments) each MDP is defined as having a randomly chosen goal state such that transitioning to it from the initial state results in a reward of +0.1, and transitioning to any other goal results in a reward of -1. Because RL2 will try to naively maximize its cumulative reward over every sub-episode within a meta-episode, it will eventually learn not to enter any goal. This is because entering an incorrect goal incurs a large negative reward of -1, a loss from which RL2 will not be able to recover by finding out which goal has +0.1 reward, whereas transitioning to the NOOP state has value 0. Therefore RL2 is not capable of producing optimal behaviour within certain classes of MDPs, especially ones where exploration has a significant risk of negative reward. A later extension, ERL2, addressed some of these concerns by splitting the meta-episode into clearly delineated exploratory and exploitative sub-episodes, and trained the policy output using only the external reward accumulated on exploit sub-episodes. This enables high-risk exploration but comes at the cost of an even more difficult temporal credit assignment problem as the external reward signal is significantly sparsified when dealing with a large number of interactions.
 
To combat the aforementioned issues that sequential meta reinforcement learning methods suffer from, we propose an alternative parallel framework termed ``Concurrent Meta Reinforcement Learning'' (CMRL). CMRL transforms the step-by-step processing of interaction episodes into a parallel processing of all episodes concurrently, reducing the meta-horizon to the time horizon of the environment. 
These parallel "rollout" agents are capable of communication with each other, enabling the assignment of specific rollouts to particular sections of the state space for efficient exploration. The communication protocol between agents therefore represents the meta-learning process, where agents transmit information about rewards, dynamics or states to each other and aggregate this information to present to the exploitation sub-episode policies in order for them to act optimally. 
By this conversion of the temporal credit assignment problem into a multi-agent communication learning problem, we are afforded other advantages beyond a reduced time horizon that no longer scales with the number of interactions. The first is the use of specialized reward sharing schemes, which can be used to re-distribute risk among the exploratory rollout policies in a way not straightforwardly possible within the sequential setting. Additionally, we can define divergence losses between rollout policies to encourage exploration and diversity of agent behaviours. 

\begin{figure}
    \centering
    \includegraphics[width=0.9\linewidth]{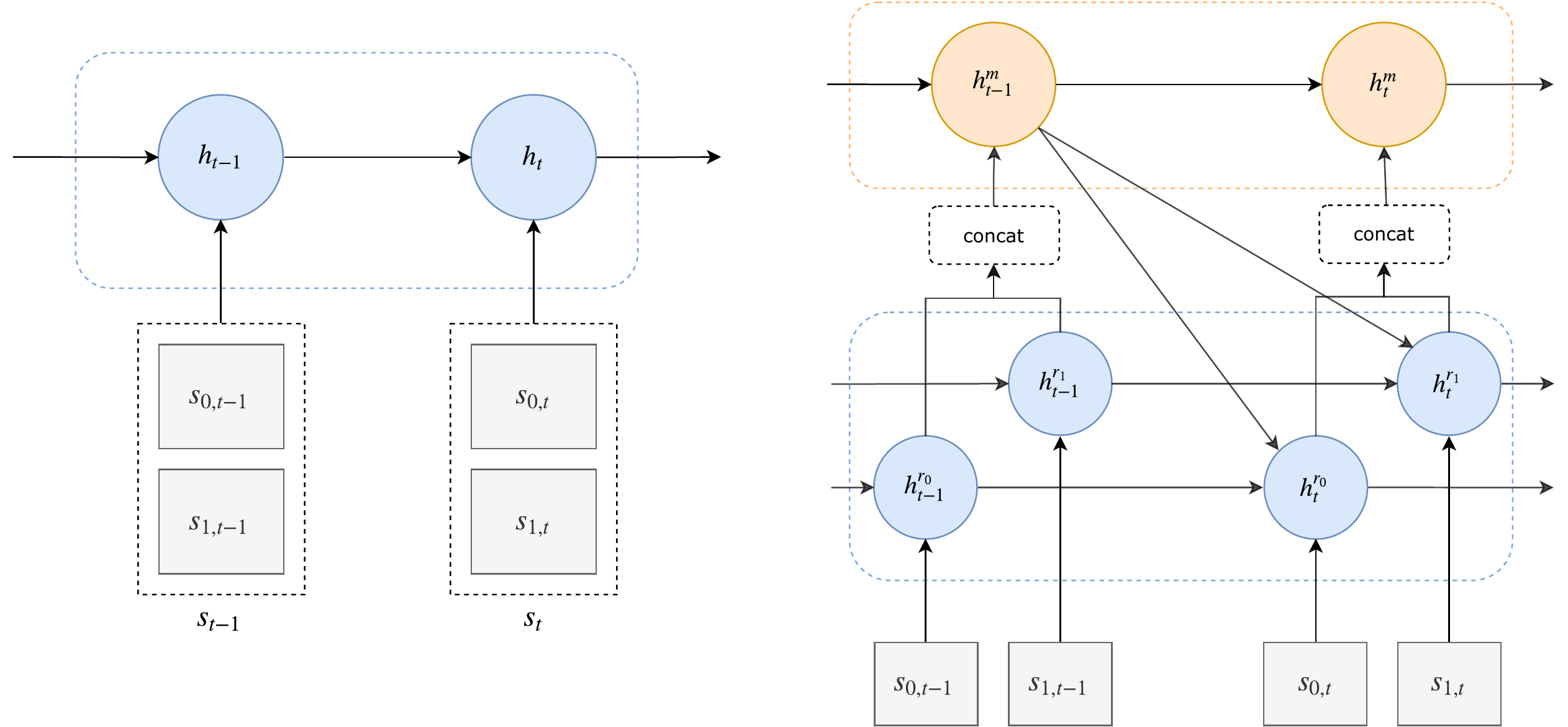}
    \caption{The two communication architectures we considered in this paper. The first is a monolithic LSTM called Central-LSTM that treats the problem as a single agent task with a factored output distribution, concatenating all per-rollout information into one combinatorial input space. The second is a more structured memory-based communication architecture called Meta-LSTM, where each rollout has a separate LSTM (with weights shared between rollouts) with a unique per-rollout learnable initial hidden state. The rollouts communicate through the Meta-LSTM, which jointly aggregates and organizes information from all rollouts.}
    \label{fig:archs}
\end{figure}

\subsection{Communication Architecture}
\label{sec:archs}

Recent work has seen a large variety of architectures developed to enable communication between agents in the setting of deep multiagent reinforcement learning~\cite{sukhbaatar2016learning, foerster2016learning, hoshen2017vain}. These various methods allow communication between agents using either discrete protocols~\cite{foerster2016learning} or continuous ones~\cite{sukhbaatar2016learning, foerster2016learning}, and they have been shown to be effective at solving  multiagent tasks. While CRML does not prescribe a particular method of inter-rollout communication, we detail here the architectures utilized in the experiments, which are illustrated in Figure~\ref{fig:archs}. 

{\bf Central-LSTM}: The first communication architecture considered is the simplest, a centralized LSTM which aggregates all rollout information (observations, rewards, terminal flags, etc.) into a single monolithic input. The Central-LSTM thus processes the multiagent data as if it were from a single-agent task with a combinatorial input space and a factored output distribution. The notion of communication within Central-LSTM is therefore absolute as every feature in the network has access to all rollout information and there are no clear delineations between tasks. Due to the combinatorial nature of Central-LSTM, learning can potentially be more challenging as there are no inductive biases built within its communication structure. 

{\bf Meta-LSTM}: The second communication architecture we consider has more structure suited to the particular application of CMRL. In this architecture, each rollout has a separate ``Rollout-LSTM'' with shared recurrent and input weights. Communication occurs through a stateful communication channel represented by a centralized ``Meta-LSTM'' that aggregates and processes information from all the separate Rollout-LSTMs, and then sends its features on the next time-step to each Rollout-LSTM. The high-level structure of this architecture is shown on the right-side of Figure~\ref{fig:archs}.

In more detail, at every time-step each Rollout-LSTM takes as input both $x_{k,t}$ (e.g. the external information from the environment such as observation, reward, terminal flag, etc.), and the previous time-step Meta-LSTM hidden state. Afterwards, the Meta-LSTM takes all the just calculated current time-step Rollout-LSTM hidden states as input to update its own hidden state. This process is represented by the update equations listed below:
\begin{align*}
    \mathbf{h}^{r}_{k,t} &= LSTM_{rollout}([\mathbf{x}_{k,t}; \mathbf{h}^{meta}_{t-1}], \mathbf{h}^{r}_{k,t-1}) \\
    \mathbf{h}^{meta}_{t} &= LSTM_{meta}([\mathbf{h}^{r}_{0,t}; \dots; \mathbf{h}^{r}_{K-1,t}], \mathbf{h}^{meta}_{t-1}) \\ 
\end{align*}
where $k$ is the index of the rollout, $t$ is the current time-step and for notational simplicity we omit the LSTM cell states from the equations. The initial hidden states $\{\mathbf{h}^{r}_{k,0}\}_{k=0}^{K-1}$ and $\mathbf{h}^{meta}_0$ are learnable parameters of the architectures.

\subsection{Reward Sharing Schemes}

Parallel rollouts allow the use of a variety of reward sharing schemes. An effective reward sharing scheme is one which can redistribute risk amongst the rollouts but still allow external signals to shape the training of the policy during exploration sub-episodes. This is in contrast to the scheme used in ERL2 which completely disregards exploratory rewards during policy optimization, sparsifying the reward signal. CMRL affords more flexibility in the design of these schemes than the sequential framework, because in the sequential case there is a significant asymmetry in the exploratory sub-episodes (sub-episode 3 is conditioned on more information than 1 and 2, etc.).  We detail a subset of possible schemes below: 

{\bf Separate Rewards}: The use of separate rewards is similar to the original RL2 reward function, where each rollout is trained to maximize its own reward regardless of whether it is an exploratory or exploitative sub-episode. This can cause competition between the rollouts, because if an exploratory behaviour requires taking a loss then all rollouts will avoid it, potentially at the cost of sub-optimal behaviour within the exploit sub-episode.

{\bf Shared Reward:} The shared reward is the common style of reward often used in collaborative multiagent reinforcement learning, where each rollout receives an external reward signal which is the summed reward of all rollouts. Because, like "Separate Rewards", no distinction is made between exploration and exploitation, Shared Rewards can suffer from the same pitfalls as RL2.

{\bf Zero-Until-Exploit:} This mirrors the ERL2 reward scheme, where rewards obtained in exploratory sub-episodes do not contribute to the training of the policy networks, and the network is trained only to optimize performance in exploitation sub-episodes.

{\bf Max-Until-Exploit:} Given $K_{explore}$ rollouts executing and communicating in parallel, the Max-Until-Exploit reward takes as external reward signal the maximum over all rollouts during an exploration sub-episode, and the raw per-rollout reward during exploit sub-episodes. Thus during exploration sub-episodes, the Max-Until-Exploit scheme will cause agents to try to maximize at least one rollout's return, while the other rollouts are free to execute high-risk exploratory behaviours like ERL2/Zero-Until-Exploit. This mixes the benefit of ERL2's risk-robustness while still enabling an external signal on the performance of the exploration sub-episodes.

{\bf StDev-Until-Exploit:} The Standard Deviation(StDev)-Until-Exploit scheme takes as reward the standard deviation of all rollout external rewards as the reward during explore sub-episodes, and the per-rollout rewards during exploit sub-episodes. The StDev-Until-Exploit scheme therefore gets higher reward when all agents produce behaviours which result in a variety of returns. While this reward sharing scheme will not be aligned to every task's objective, we found it to provide some useful learning signal in very high-risk environments.

In this paper, we consider the two risk-robust schemes Max-Until-Exploit and StDev-Until-Exploit for training the concurrent agents, which allow an external learning signal to be taken into consideration while still enabling rollouts to take high-risk exploratory behaviours during exploration sub-episodes.

\subsection{Divergence Losses}

Beyond reward sharing schemes, we develop several auxiliary losses which can help produce varied behaviours in the population of agents.
The goal of these auxiliary losses is to ideally drive agents into different partitions of the state space, to encourage exploration to new areas of the environments. Given that we potentially want to deal with partially-observable environments, and that even in fully-observable environments we require memory between episodes, we develop a framework that calculates divergences while still maintaining the proper memory states of each rollout.

We first describe the setting in which auxiliary losses are defined. Let $K$ be the number of rollout agents executing in parallel, and let $(s_1^{(i)},\ldots,s_H^{(i)})$ and $(a_1^{(i)},\ldots,a_H^{(i)})$ define parallel rollouts agent $i$'s experienced state and action sequence, respectively, which could potentially cover several explore sub-episodes. Let $h_t^{(i)}(s_1^{(j)},a_1^{(j)},\ldots,s_t^{(j)})$ be the memory state of agent $i$ at time-step $t$, when agent $i$ is run on the history of states $(s_1^{(j)},\ldots,s_t^{(j)})$ collected by (a potentially different) agent $j$. Let $\pi^{(i)}(h_t)$ be the policy of agent $i$ computed from hidden state $h_t$. We define the auxiliary divergence losses in the following form: 
\begin{align*}
  L_D = \sum_{i,j} \sum_{t=0}^H D(\pi^{(i)}(h_t^{(i)}(s_1^{(i)},\ldots,s_t^{(i)})), \pi^{(j)}(h_t^{(j)}(s_1^{(i)},\ldots,s_t^{(i)})))
\end{align*}
This loss encourages different rollout policies to output distinct action probabilities on the same state history input, thereby achieving more diverse behaviours. Given the ${K \choose 2}$ possible pairs of states we could consider, we implement this efficiently in the following manner. We first cache the values of $\pi^{(i)}(h_t^{(i)}(s_1^{(i)},\ldots,s_t^{(i)}))$ obtained during the actual policy rollout. We then sample a permutation $P$ and rollout all $K$ agents simultaneously on the permutation's state and calculate $L_D$. To further improve efficiency considering we already obtained the values of the rollout policies run on their own state during experience collection, we restrict the permutations we sample to derangements (where no element appears in its original position), with the number of derangements sampled being a hyperparameter.

In our experiments, we consider two forms of symmetric divergences $D(\pi^{(i)}, \pi^{(j)})$. The first is the symmetric {\bf Kullback-Leibler Divergence}:
$D_{SKL}(\pi^{(i)}, \pi^{(j)}) = D_{KL}(\pi^{(i)}||\pi^{(j)}) + D_{KL}(\pi^{(j)}||\pi^{(i)})$
The second is the {\bf Jensen-Shannon Divergence}:
$D_{JS}(\pi^{(i)},\pi^{(j)}) = \frac{1}{2} D_{KL}(\pi^{(i)}||\pi^M) + \frac{1}{2} D_{KL}(\pi^{(j)}||\pi_M)$, 
with $\pi^M = \frac{1}{2} \left(\pi^{(i)} + \pi^{(j)}\right)$. The divergence loss is modulated with respect to the RL loss with the use of a hyperparameter $\lambda_{div}$.

\section{Experiments}

\begin{figure}
    \centering
    \includegraphics[width=0.3\linewidth]{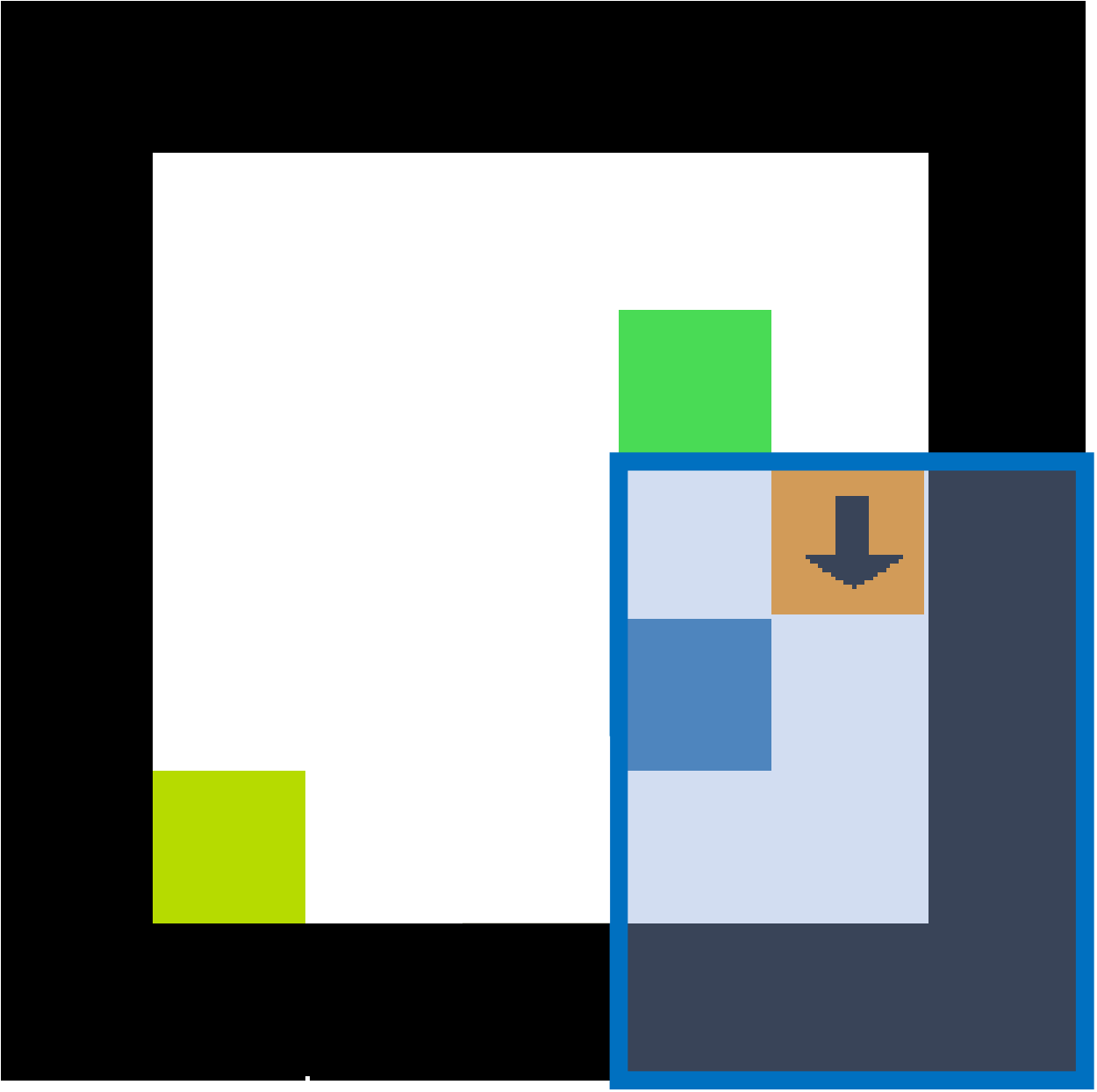} $\qquad$
    \includegraphics[width=0.3\linewidth]{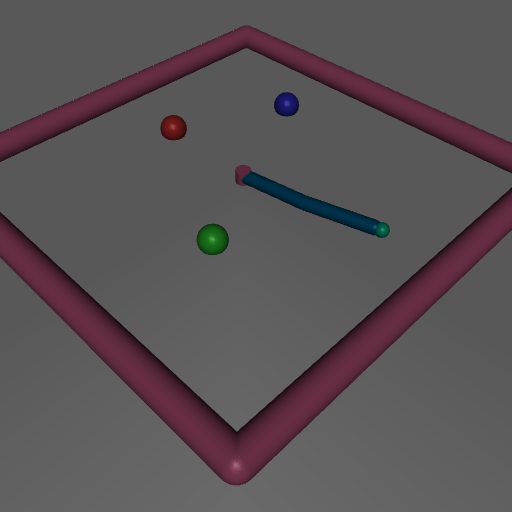}
    \caption{{\bf Left:} An example of a $7\times 7$ 3-Color-Choice environment with the 3 goals shown as yellow, green and blue. The agent is depicted by the orange square with an arrow showing its current orientation. The agent's current field of view, which represents the pixels it is observing as input to its policy, is shown in the transparent blue rectangle. The goal of the environment is to navigate the room, and determine which of the colored goals contains the positive reward. {\bf Right:} An example of the 3-Reacher environment, with two target end-effector positions shown in red and green. The agent does not know which position has the positive reward beforehand and must query the goal states during its exploration phase. }
    \label{fig:visenv}
\end{figure}

\begin{figure*}
  \centering
  \includegraphics[width=0.338\linewidth,valign=t]{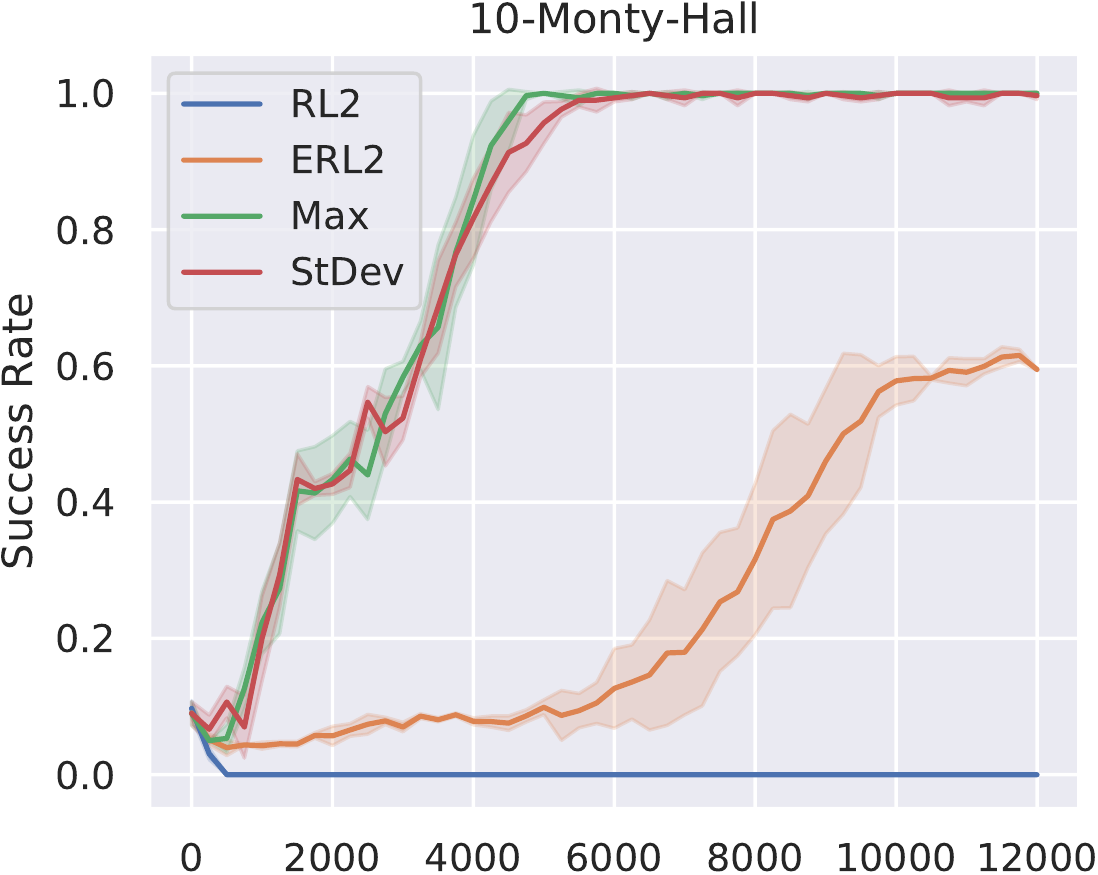}%
  \includegraphics[width=0.33\linewidth,valign=t]{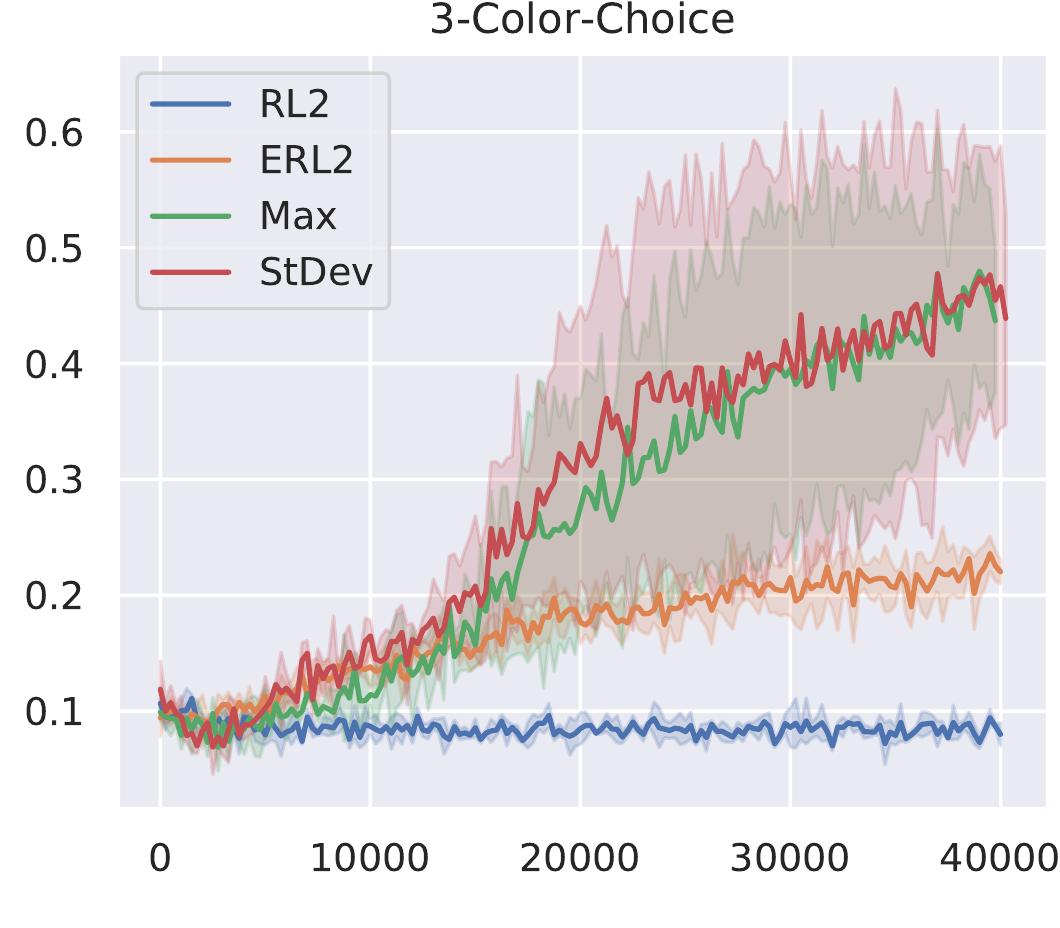}%
  \includegraphics[width=0.33\linewidth,valign=t]{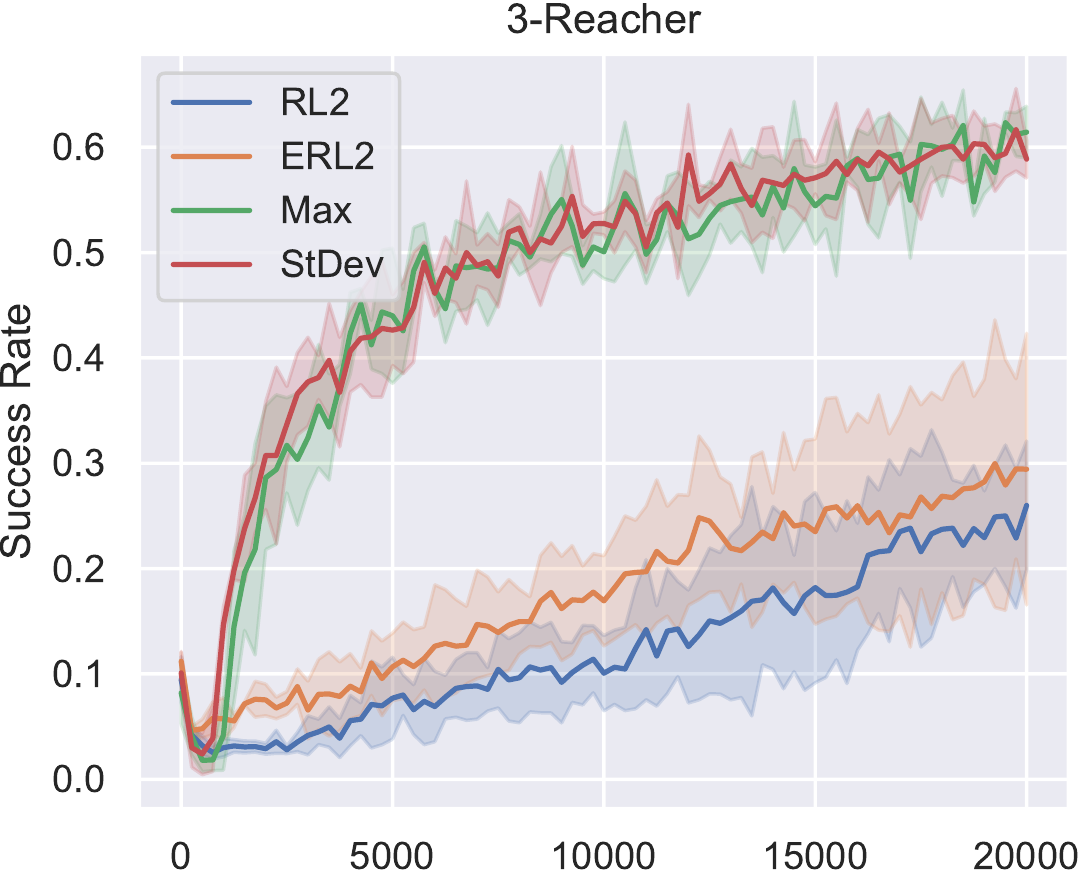} \\
  \mbox{
  \includegraphics[width=0.335\linewidth,valign=t]{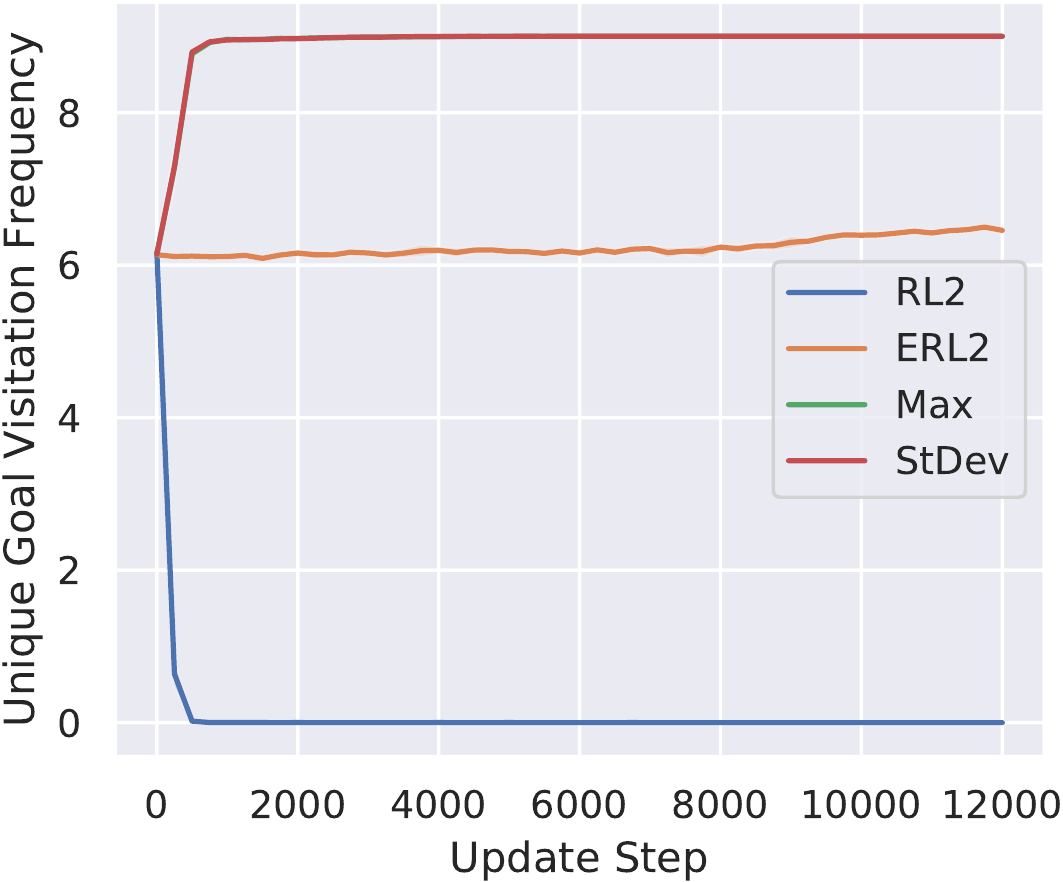}%
  \includegraphics[width=0.33\linewidth,valign=t]{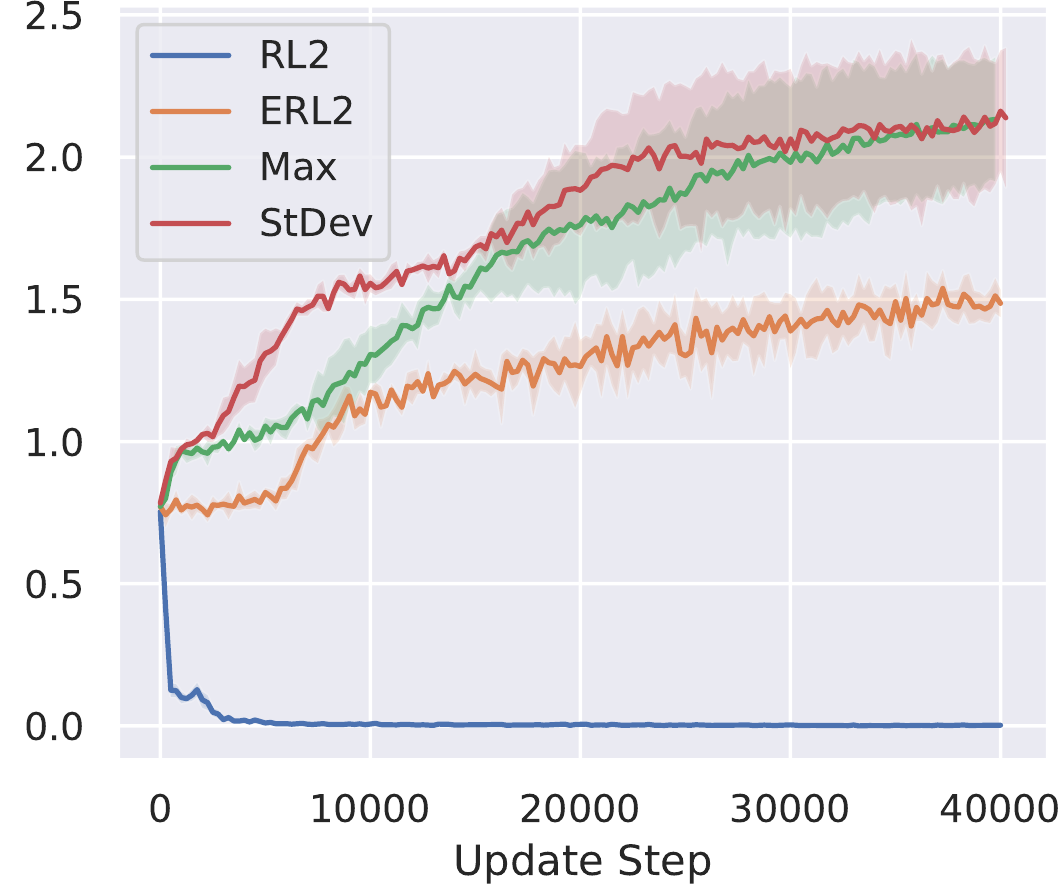}
  \includegraphics[width=0.33\linewidth,valign=t]{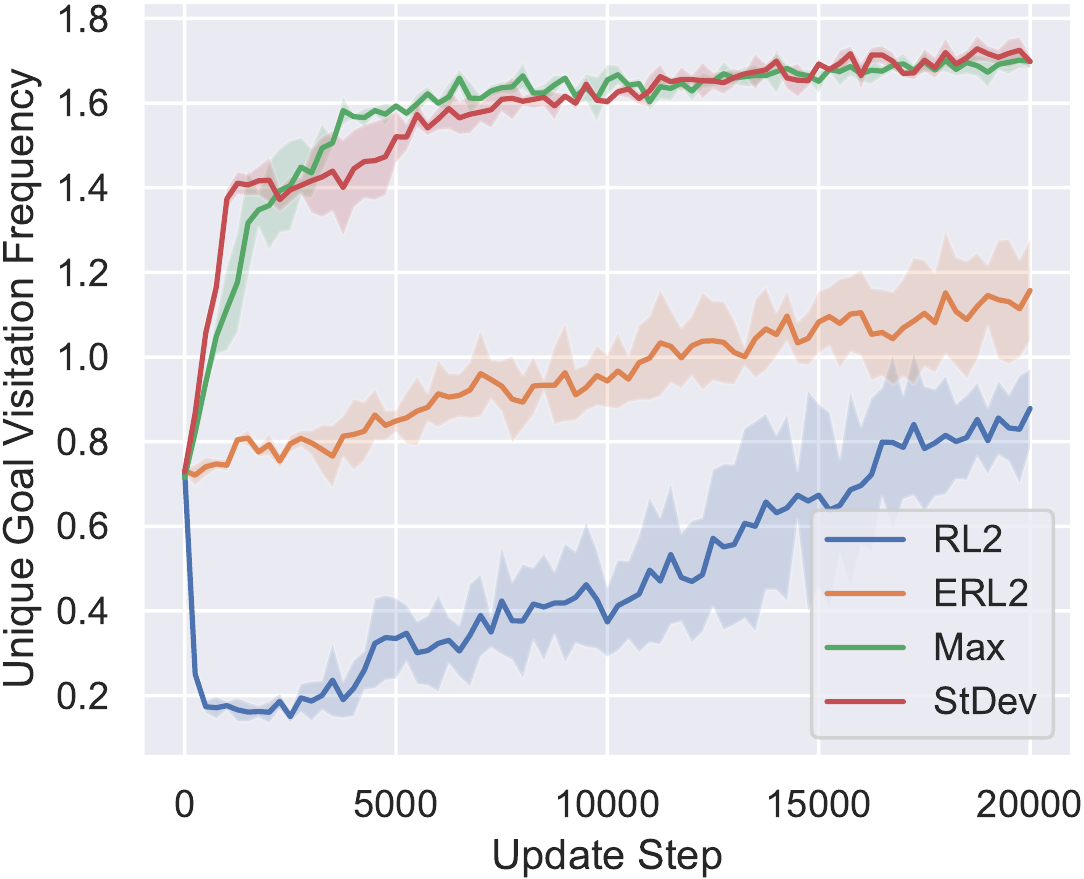}
  }
  \caption{Learning curves of the four models on the three environments: 10-Monty-Hall (Left), 3-Color-Choice (Center), 3-Reacher (Right). The top row shows the success rate of the model over epochs. The bottom row shows unique goals visitation frequency during explore sub-episodes. To obtain each datapoint in the curve, we average results over 1000 meta-episodes and 3 seeds per model.}
    \label{fig:learning}
\end{figure*}

We validate our concurrent meta-learning method on three environments, comparing against the baseline episodic meta-RL algorithms RL2/L2RL (referred to as RL2 in the sequel for simplicity) and ERL2. The first class of environments we consider are the $N$-Monty-Hall MDPs described at the start of Section~\ref{sec:cmrl} but generalized to $N$ doors. These environments provide a setting that clearly demonstrates empirically that RL2 is highly susceptible to failure in high-risk environments and that ERL2 can receive too little external reward to learn efficiently. We then consider a class of challenging partially-observable navigation tasks called $N$-Color-Choice, where several colored goals are randomly located within a grid-world-like environment. The agent can view only a subset of the environment at once and must discover which of the goals has positive reward (every other having negative reward). The third class of environments, $N$-Reacher, is a discretized version of the standard Reacher-v2 robotics environment commonly used as a baseline for continuous control algorithms, except there are multiple target locations spread out within the environment and the agent must learn to reach one of the targets while avoiding all others, without knowing a priori which of the targets it should hit. We now describe each environment in detail:

{\bf $\textbf{N}$-Monty-Hall}: The ``Monty-Hall'' environment (depicted visually in Figure~\ref{fig:unsolvable}) is the class of MDPs described in Section~\ref{sec:cmrl}. Each episode in the environment consists of an agent choosing from a set of $N$+1 actions. The first action is a NOOP, where the agent does not receive reward. The other $N$ actions choose from a set of doors, behind which is an unobservable reward not known a-priori to the agent. For a given MDP from this class, only one fixed door will always give a small positive reward of +0.1 (represented by door which leads to the gold) while all others give large negative rewards of -1 (represented by the doors leading to bombs). Succesful exploration in this environment therefore needs the agent to potentially accumulate large negative reward in order to eventually find the door leading to the gold. This setting is meant to demonstrate the issues with RL2's training regime, which takes into consideration external exploration sub-episode rewards and tends to produce very risk-adverse meta-learners. 

{\bf $\textbf{N}$-Color-Choice}:
$N$-Color-Choice is a navigation environment, with $N$ target absorbing goals $\{G_i\}_{i=1}^N$ randomly spawned within a $H_M \times W_M$ grid.
An agent starts within the $H_M\times W_M$ grid facing a random orientation. This orientation is sampled from the discrete set of cardinal directions $\{$North, East, West, South$\}$. The transition dynamics of the agent are described by a differential drive, meaning there are 3 possible actions: it can move forward to a neighboring un-occupied grid cell along its cardinal orientation, or it can either turn left or right to change its orientation. Moving forward into a wall does not change the agent position. The agent views objects within the grid cells 15 pixels ahead of itself with respect to its current orientation and 1 pixel on each side of this ray, resulting in an observation input image of dimension $C\times 15\times 3$, where the number of channels $C$ is dependent on the semantics of the particular environment.
Each goal in the environment is assigned a unique color which is observable by the agent's sensors. That is, for each goal $G_i$ there exists a separate channel in the $C\times 15\times 3$ observation, i.e. $c(G_i)$. If $G_i$ is observed at observation cell $(y,x)$ relative to the player's current position and orientation, than channel $c(G_i)$ is 1 at position $(y,x)$, with it being 0 otherwise. Furthermore, each goal $G_i$ is assigned a hidden reward $r(G_i)\in\{-1,1\}$ which is externally un-observable by the observation sensors within an episode, meaning there is no possibility to determine $r(G_i)$ using the observation history within a single episode. The rewards are set such that only one unique goal has positive reward +1, i.e. there exists some $G_j$  with $r(G_j) = +1$ and $r(G_i) = -1$ for all $i\neq j$. The goal of the meta-learner is thus to learn to enter each of the coloured goals and remember which color had the positive reward. A visualization of the fully observable environment state and agent observation are shown in Figure~\ref{fig:visenv}. 

{\bf $\textbf{N}$-Reacher}:
$N$-Reacher is a variant of the $N$-Color-Choice environment but extended to a Mujoco manipulation task. Built off the mujoco benchmark environment ``Reacher-v2'', this environment extends that setting to have multiple target positions and discretizes actions to choosing torques of $\{-1,0,1\}$ at each time-step. Instead of an agent navigating a grid towards a goal cell, $N$-Reacher has a 2-joint arm that must position its end-effector near a target position. The episode ends whenever the agent gets below a distance threshold to the target. As in $N$-Color-Choice, there are $N$ targets where only one of the targets has +1 and the others have -1. Each MDP in the class has target positions varying within the $[-1,1]\times [-1,1]$-sized area and random reward assignments. An example rendering of the environment is shown on the right of Figure~\ref{fig:visenv}.

In each environment, we set a hard-limit time horizon of $H$ steps per sub-episode, meaning that after $H$ action choices by the agent we terminate the episode whether the agent has already entered an absorbing goal state or not. Technically we require $N-1$ exploratory sub-episodes in order to solve the environments but for simplicity we give each of the meta-agents $N$ exploratory sub-episodes (i.e. $k_{explore} = N$) and $1$ exploitative sub-episode (i.e. $k_{exploit} = 1$). All sub-episodes for a particular meta-episode have the same initial state. The MDPs are randomly sampled each meta-episode. We run experiments for $10$-Monty-Hall, $3$-ColorChoice and $3$-Reacher.

For each environment we train baselines RL2 and ERL2 as well as our proposed CMRL. The types of communication architectures we used for CMRL are as described in Section~\ref{sec:archs}, and the particular one used depended on the environment. For Monty-Hall we used CentralLSTM while for Color-Choice and Reacher we used MetaLSTM. All models were trained using A2C with a discount factor of $0.99$ and the Adam optimizer. We set $H$ to 1 for Monty-Hall and 15 for the other environments. The best models based on training curves were chosen. Models were trained for 10k steps for Monty-Hall, 40k for Color-Choice and 20k for Reacher. More architecture details are given in Appendix~\ref{sec:hyperparam}.

For evaluations, we checkpointed model files every 250 update steps and ran 10 batches of 128 samples (1280 meta-episodes in total) to obtain learning curves and final results. We report a variety of statistics: ``AuC'' (Area-under-Curve) is the mean area under the learning curves and gives a quantitative notion of learning speed, ``Final Perf'' is the success rate of the meta-learner after training, Updates until Success Rate represent the number of update steps until mean success rate is $\geq$ a specified threshold, ``Visited Goals'' is the number of unique goals (different absorbing states) visited during exploration sub-episodes.

\begin{table*}[t]
  \centering
  \begin{tabular}{@{}llrrrrrrr@{}}
    \multirow{2}{*}{Model} & \multirow{2}{*}{AuC} & \multirow{2}{*}{Final Perf.} & \multicolumn{5}{c}{Updates until Success Rate} & \makecell{Visited \\ Goals} \\
    & & & 25\% & 50\% & 75\% & 95\%& 100\% & \\
    \hline
    RL2          & 19.83 & 0.00 {\small $\pm$ 0.0\%} & - & - & - & - & - & 0.00 {\small $\pm$ 0.0} \\
    ERL2         & 2939.17 & 59.5 {\small $\pm$ 0.1\%} & 7500 & 9250 & - & - & - & 6.42 {\small $\pm$ 0.0} \\
    \hline
    Max-Until-Exploit & {\bf 9575.42} & {\bf 100 {\small $\pm$ 0.0\%}} & 1250 & 2750 & 3750 & 4500 & 5000 & {\bf 9.00 {\small $\pm$ 0.0}} \\
    StDev-Until-Exploit & {\bf 9501.67} & {\bf 99.7 {\small $\pm$ 0.5\%}} & 1250 & 2500 & 3750 & 5000 & 6500 & {\bf 9.00 {\small $\pm$ 0.0}} \\
    \hline
  \end{tabular}
  \caption{Table showing results on the 10-Monty-Hall environment. See text for details. }
  \label{tab:mhresults}
\end{table*}

\begin{table*}[t]
  \centering
  \begin{tabular}{@{}llrrrrrr@{}}
    \multirow{2}{*}{Model} & \multirow{2}{*}{AuC} & \multirow{2}{*}{Final Perf.} & \multicolumn{4}{c}{Updates until Success Rate} & \makecell{Visited \\ Goals} \\
    & & & 10\% & 20\% & 40\% & 100\% & \\
    \hline
    RL2 & 3408.67 & 8.0 {\small $\pm$ 1.0\%} & 0 & - & - & - & 0.00 {\small $\pm$ 0.0} \\
    ERL2 & 6811.58 & 22.0 {\small $\pm$ 1.0\%} & 2750 & 23750 & - & - & 1.49 {\small $\pm$ 0.1} \\
    \hline
    Max-Until-Exploit & {\bf 10374.25} & {\bf 43.7 {\small $\pm$ 5.0\%}} & 5000 & 15750 & 30750 & - & {\bf 2.13 {\small $\pm$ 0.2}} \\
    StDev-Until-Exploit & {\bf 11436.83} & {\bf 43.9 {\small $\pm$ 7.5\%}} & 0 & 14500 & 28000 & - & {\bf 2.14 {\small $\pm$ 0.3}} \\
    \hline
  \end{tabular}
  \caption{Table showing results on the 3-Color-Choice environment. See text for details.}
  \label{tab:ccresults}
\end{table*}

\begin{table*}[t!]
  \centering
  \begin{tabular}{@{}llrrrrrr@{}}
    \multirow{2}{*}{Model} & \multirow{2}{*}{AuC} & \multirow{2}{*}{Final Perf.} & \multicolumn{4}{c}{Updates until Success Rate} & \makecell{Visited \\ Goals} \\
    & & & 10\% & 20\% & 40\% & 100\% & \\
    \hline
    RL2 & 2500.25 & 26.0 {\small $\pm$  5.0\% } & 7500 & 16250 & - & -& 0.87 {\small $\pm$ 0.09} \\
    ERL2 & 3521.38 & 29.4 {\small $\pm$  10.5 \% } & 0 & 11250 & - & - & 1.16 {\small $\pm$ 0.12} \\
    \hline
      Max-Until-Exploit & {\bf 9399.12} & {\bf 61.4 {\small $\pm$  2.0\% }} & 1250 & 1750 & 4000 & -
 & {\bf 1.70 {\small $\pm$ 0.02}} \\
      StDev-Until-Exploit & { \bf 9626.29 } & { \bf 58.9 {\small $\pm$  1.5\% } } & 0 & 1500 & 4000 & -
 & {\bf 1.70 {\small $\pm$ 0.01}} \\
    \hline
  \end{tabular}
  \caption{Table showing results on the 3-Reacher environment. See text for details.}
  \label{tab:rresults}
  \end{table*}

{\bf 10-Monty-Hall Results:}
The results on the 10-Monty-Hall environments are presented in Table~\ref{tab:mhresults} with learning curves of the average exploit sub-episode return on the top row and goals entered during exploration sub-episodes on the bottom row of Figure~\ref{fig:learning}. We can see a significant difference between CMRL and the baselines in learning speed, final performance, and meta-learner exploration. As expected, the first row demonstrates that RL2 completely fails at this environment, with the policy rapidly collapsing onto the NOOP action as demonstrated by the learning curves in Figure~\ref{fig:learning}, converging to 0 success rate and 0 unique goals entered during exploration. In contrast, ERL2 performs better due to its policy having no pressure to avoid entering the negative goals, so it sustains an average of 6 goals entered. The difficulty is that it does not have much incentive to explore, and the external reward of the exploit episode has to propagate backwards over a longer time horizon to dictate exploratory behaviour. The unique goals entered by ERL2 is only very slightly increasing over update steps, only reaching 6.42 from an initial 6 after training for 12000 update steps.
In contrast, CMRL very rapidly converges to optimal behaviour, reaching 9 goals entered within less than 1000 update steps. We found that CMRL often preferred this 9 goals entered over 10 goals, since it learnt to delegate one of its exploration rollouts to always perform the NOOP action. This enables it to learn with relatively low risk, and it tended to converge to this policy after which exploration (at the meta-learner level) was less likely. 9 goals entered is still an optimal exploration policy because the last reward can be inferred, which is reflected in the optimal performance of the exploitaiton policy after training.
In this environment there is no statistically significant difference between the Max-Until-Exploit and StDev-Until-Exploit reward sharing schemes, with both achieving similar results with Max-Until-Exploit having a slightly higher mean learning speed and final performance over StDev-Until-Exploit.

{\bf 3-Color-Choice Results:}
The results of 3-Color-Choice are shown in Table~\ref{tab:ccresults}, with the learning curves in the center column of Figure~\ref{fig:learning}. For RL2, we can see a similar situation to 10-Monty-Hall, with a rapid collapse to no goal entries during exploration sub-episodes due to the high-risk involved. Also similar to 10-Monty-Hall, ERL2 is capable of learning but does so at a much slower pace than CMRL. We can see that during exploration, ERL2 only learns to enter about half of the available goals. Despite the 1.5/3 average goals entered during exploration sub-episodes, ERL2 still has trouble making the connection between reward of goals entered and optimal exploit policy, only reaching average 0.2 exploit return. This might be due to the much longer time horizons involved in the 3-Color-Choice setting making credit assignment more challenging. From the table, we can see CMRL learns both faster and at a higher performance than all baselines.

{\bf 3-Reacher Results:}
Results for 3-Reacher are shown in Table~\ref{tab:rresults} with learning curves on the right column of Figure~\ref{fig:learning}. The results in the table show a substantial improvement in learning speed and final performance over the baselines RL2 and ERL2. Similar to other environments, the different reward schemes achieve similar performance. RL2, in contrast to its performance in previous environments, actually demonstrated learning on this environment. The reward scales in 3-Color-Choice and 3-Reacher are high-risk but do not preclude RL2 from solving them like 10-Monty-Hall, only having a strong local minima for the NOOP policy which is then difficult to exit using exploration. For 3-Reacher, it does seem like RL2 exited this local optimum after reaching it, as seen by the goal visitation curve dipping to 0 very early on in training and then recovering afterwards at a rate much higher than ERL2. One possible reason for RL2 having difficulty in 3-Color-Choice is its partial-observability, causing learning to be substantially slower (almost two times) in 3-Color-Choice when compared to the fully-observable state of 3-Reacher, and making it more likely for RL2 to collapse to the NOOP policy after which exploration stops. For CMRL, the learning curves demonstrate that the exploit sub-episode can achieve a high score even though its exploration policy is sub-optimal (not reaching the 2 unique goals visited necessary for the exploit target goal to be determined).

\begin{figure}
  \centering
  \mbox{
    \begin{minipage}{0.33\linewidth}
      \includegraphics[width=\linewidth]{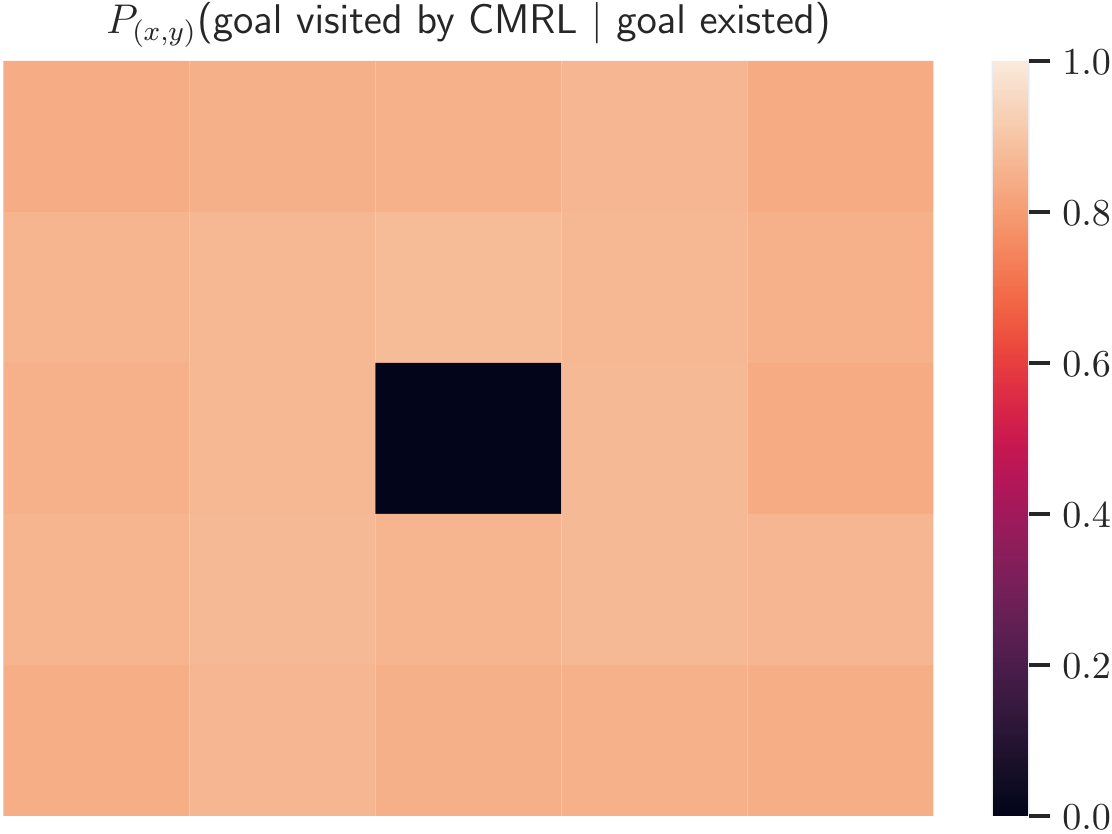}
    \end{minipage}%
    \begin{minipage}{0.01\linewidth}
      $ $
    \end{minipage}%
    \begin{minipage}{0.33\linewidth}
      \includegraphics[width=\linewidth]{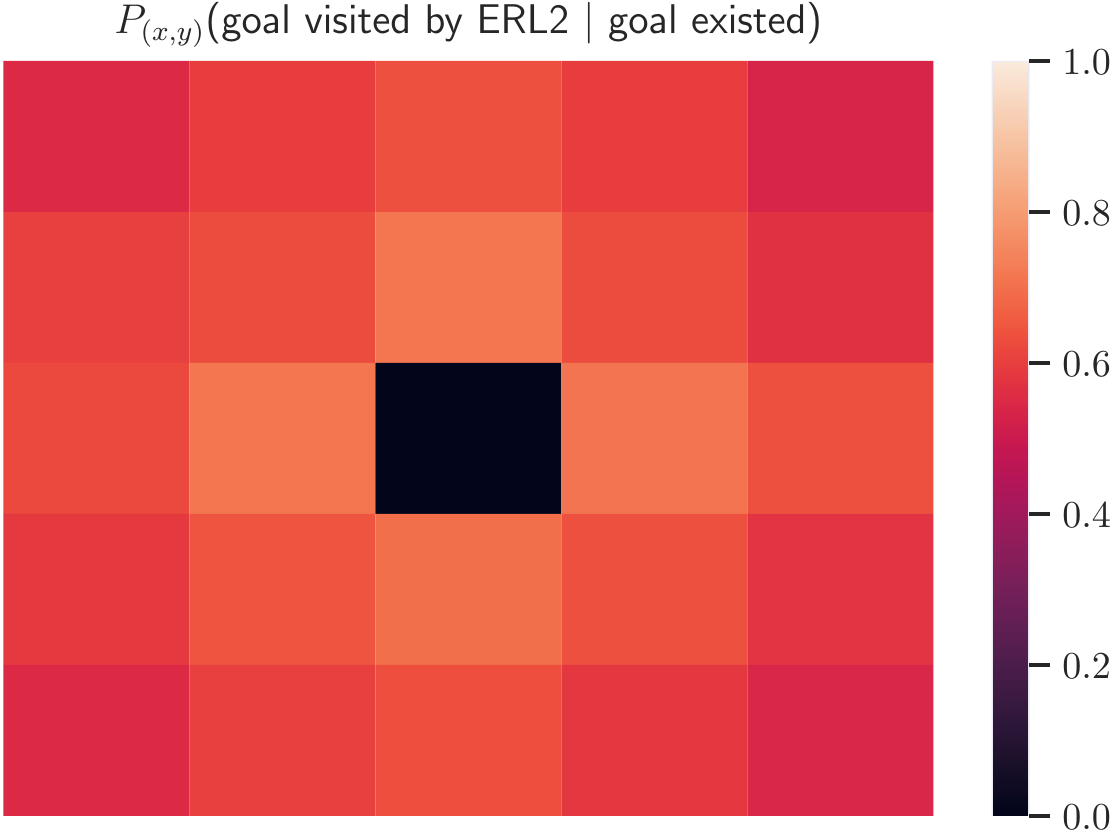}
    \end{minipage}%
    \begin{minipage}{0.01\linewidth}
      $ $
    \end{minipage}%
    \begin{minipage}{0.33\linewidth}
      \includegraphics[width=\linewidth]{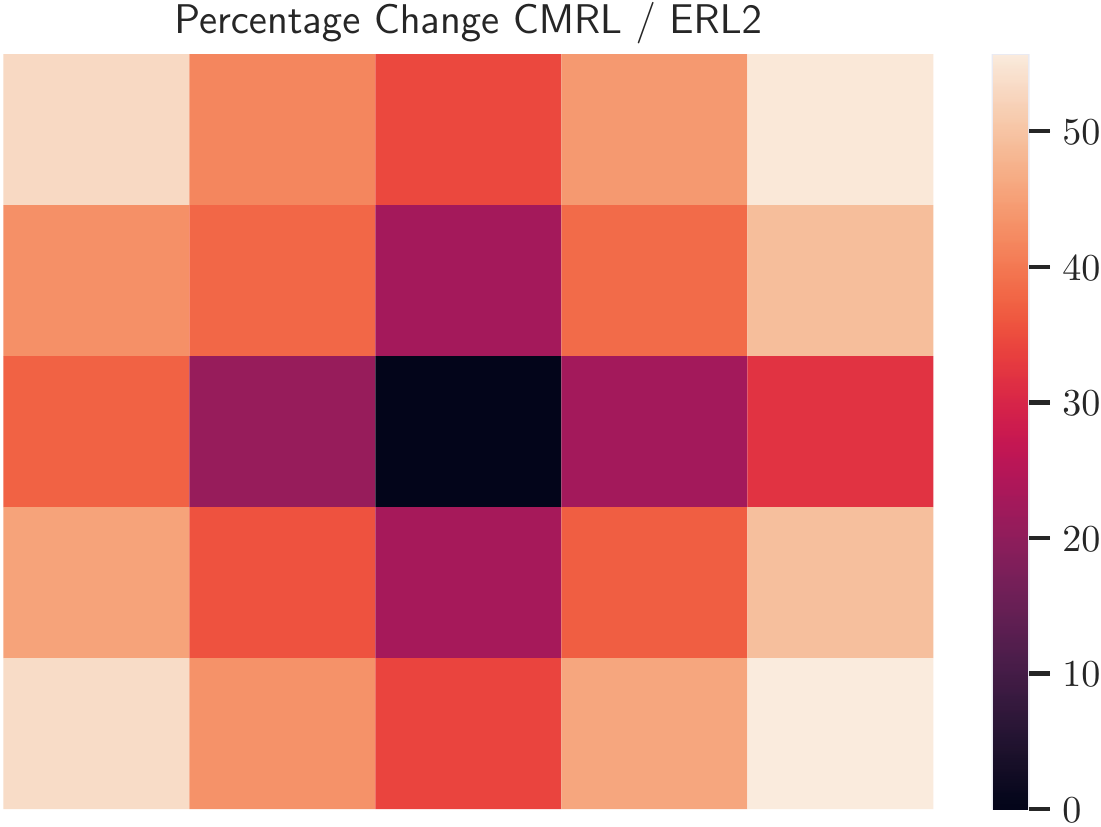}
    \end{minipage}
  } \\
  \mbox{
    \begin{minipage}{0.33\linewidth}
      \includegraphics[width=\linewidth]{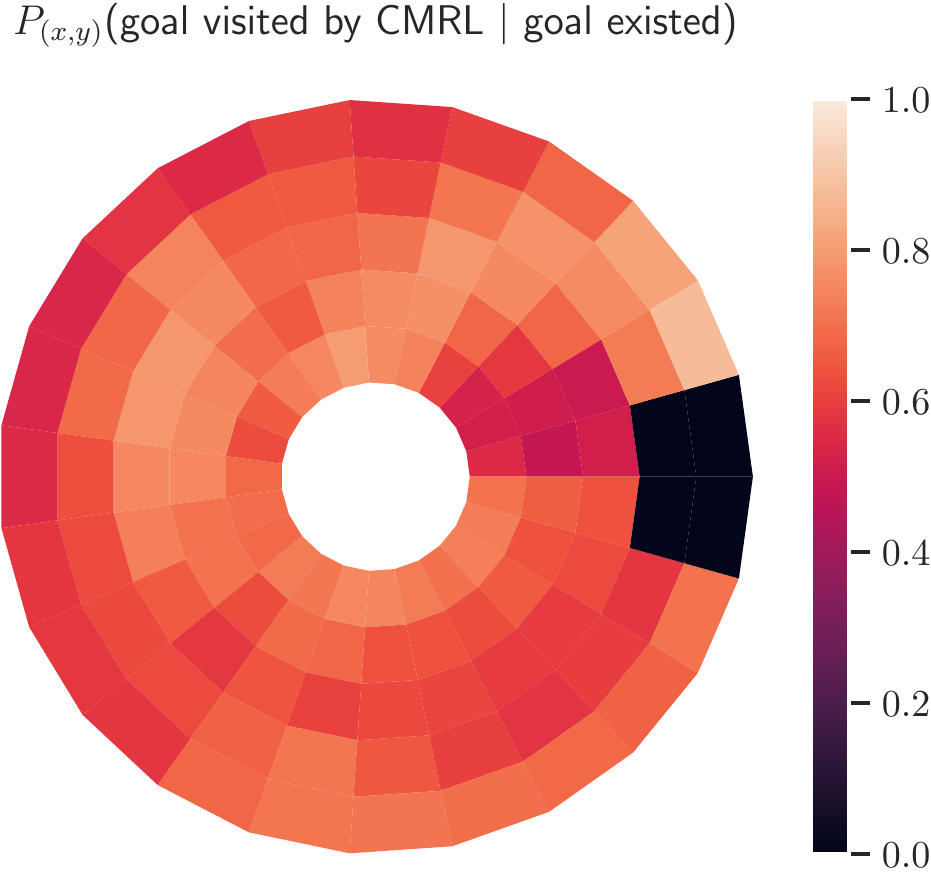}
    \end{minipage}%
    \begin{minipage}{0.01\linewidth}
      $ $
    \end{minipage}%
    \begin{minipage}{0.33\linewidth}
      \includegraphics[width=\linewidth]{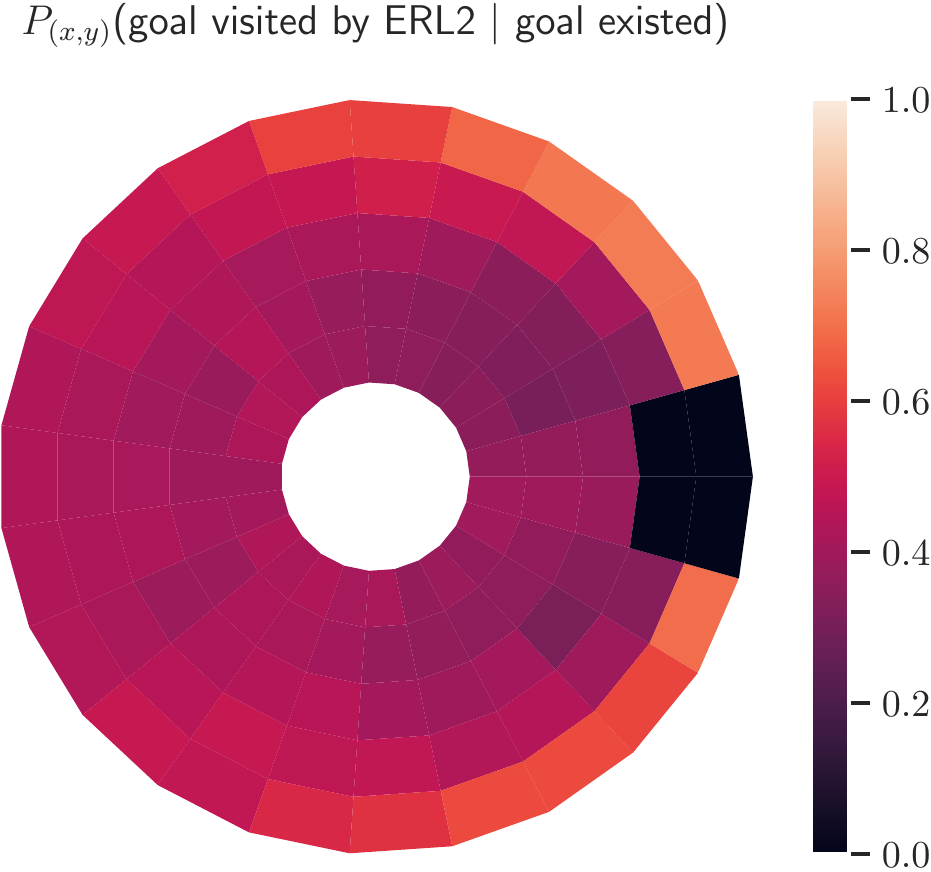}
    \end{minipage}%
    \begin{minipage}{0.01\linewidth}
      $ $
    \end{minipage}%
    \begin{minipage}{0.33\linewidth}
      \includegraphics[width=\linewidth]{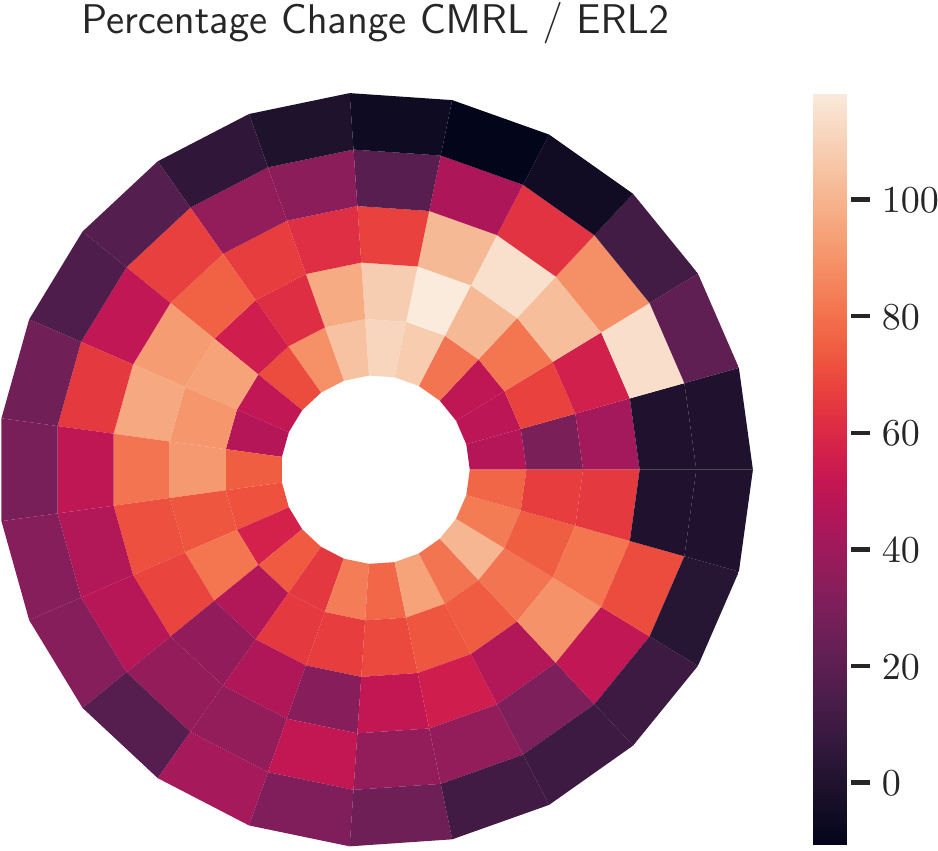}
    \end{minipage}
  }
  \caption{To examine the exploration capability of CMRL versus ERL2 in 3-Color-Choice (Top) and 3-Reacher (Bottom), we measured the amount of times CMRL explored a particular $(x,y)$ position given that the goal was located at $(x,y)$ over 40000 random meta-episodes. Compared to ERL2, we found a substantial increase in not only the number of unique goals visited during exploration sub-episodes as reported in Table~\ref{tab:ccresults}~\ref{tab:rresults} but also their variety. 3-Reacher goals were quantized in the visualizations due to continuous positions. {\bf Left:} CMRL's visitation frequency for every possible goal location. {\bf Center:} ERL2's visitation frequency for every possible goal location. {\bf Right:} Percent change from ERL2 to CMRL of visitation frequency for every possible goal location.}
  \label{fig:visitvis}
\end{figure}

{\bf Improved Exploration of CMRL:}
We also examined the final exploratory behaviour of the CMRL and ERL2 meta-learners after training in the 3-Color-Choice and 3-Reacher environment. To do so, we sample 40000 meta-episodes from the class of environments and, for each of these meta-episodes, keep track of wherever the goals were sampled and whether the agent entered this goal during exploration sub-episodes. This gives us an idea of, when a particular goal is located at position $(x,y)$, what is the frequency that the meta-learner actually explores there. We plotted this information in Figure~\ref{fig:visitvis} for both CMRL and ERL2 in the left and center columns, respectively. We can clearly notice that CMRL gets a much higher coverage of goal locations, meaning it does not only improve upon ERL2 in the number of unique goals visited during exploration sub-episodes, but the variety of such goals as well. We show this even more clearly in the heatmaps on the right column of Figure~\ref{fig:visitvis}, which shows the percent change in visitation frequency from ERL2 to CMRL for each goal location. 

\begin{table}[t]
  \centering
  \begin{tabular}{@{}llrrrrrrr@{}}
    \multirow{2}{*}{Model} & \multirow{2}{*}{AuC} & \multirow{2}{*}{Final Perf.} & \multicolumn{5}{c}{Updates until Success Rate} & \makecell{Visited \\ Goals} \\
    & & & 25\% & 50\% & 75\% & 95\%& 100\% & \\
    \hline
    {\bf CMRL} & & & & & & & & \\
    {\small Max-Until-Exploit} & {\bf 9575.96} & {\bf 100 {\small $\pm$ 0.0\%}} & 1250 & 2750 & 3750 & 4500 & 5000 & {\bf 9.00} \\
    {\small StDev-Until-Exploit} & {\bf 9501.67} & 99.7 {\small $\pm$ 0.5\%} & 1250 & 2500 & 3750 & 5000 & 6500 & {\bf 9.00} \\
    \hline
    {\bf Mixing Schemes} & & & & & & & & \\
    {\small Max+StDev-Until-Exploit} & 9448.75 & 99.0 {\small $\pm$ 0.8\%} & 1250 & 2750 & 4000 & 5000 & 6500 & {\bf 9.00} \\
    \hline
    {\bf No Reward Scheme} & & & & & & & & \\
    {\small Zero-Until-Exploit} &  9417.92 & 92.0 {\small $\pm$ 11\%} & 750 & 1750 & 3000 & - & - & 8.67 \\
    \hline
    {\bf No Divergence} & & & & & & & & \\
    {\small Max-Until-Exploit} & 51.67 & 4.67 {\small $\pm$ 6.6\%} & - & - & - & - & -  & 9.00 \\
    {\small StDev-Until-Exploit} & 45.83 & 0.0 {\small $\pm$ 0.0\%} & - & - & - & - & - & 4.64 \\
    {\small Zero-Until-Exploit}  &  40.83 & 0.0 {\small $\pm$ 0.0\%} & - & - & - & - & - & 6.15 \\
    \hline
  \end{tabular}
  \caption{CMRL ablation results for 10-Monty-Hall. We examined the contribution of mixing reward schemes, using no reward scheme, and removing the divergence auxiliary objectives. At the top is the final CMRL results from Table~\ref{tab:mhresults}. We can see that mixing reward schemes gives around the same performance as either Max- or StDev-Until-Exploit. Using no reward scheme and divergence losses does much better than baselines in final performance and learning speed, but ends up with a lower final performance than the full CMRL. Removing the divergence loss reduces learning capability in 10-Monty-Hall but interestingly Max-Until-Exploit still learns an optimal exploration policy.}
  \label{tab:ablation}
\end{table}

{\bf Ablation of the Various CMRL Components:}
We carried out an ablation on 10-Monty-Hall to try to extract out the contribution of each component of our proposed CMRL with results shown in Table~\ref{tab:ablation}. The first component we adjusted was the reward scheme. Given that the reward scheme adds an extra hyperparameter, we wanted to examine whether it was possible to just mix together a variety of reward schemes and get similar results, to avoid an additional hyperparameter to tune. This was the row labeled ``Max+StDev-Until-Exploit'', a scheme which simply sums the Max- and StDev-Until-Exploit reward schemes together during exporation sub-episodes. The result demonstrates that we can nearly recover the performance of either reward scheme using this mixture, demonstrating the robustness of the hyperparameter and suggesting that it can reasonably be set to a mixture of reward schemes. The next component we changed was examining what happens when we remove all reward schemes, making CMRL rely only on external exploit returns as ERL2 does. In this case, we can still see a substantial improvement over ERL2, which is (as explained later) mainly due to divergence auxiliary objective that encourages exploration. Despite the improvement over ERL2, Zero-Until-Exploit CMRL still performs worse and with higher variance than if we had used a reward-sharing scheme. Finally, we examined what happens to CMRL without the divergence auxiliary loss. In this case, we see a substantial decrease in exploit performance, with final performance near 0. Interestingly, the Max-Until-Exploit CMRL eventually achieves an optimal explore policy despite the poor exploit performance. This suggests that higher exploit performance is likely achievable with a finer tuning of the entropy scale parameter used in A2C, to prevent the exploit sub-episode policy from collapsing ``too early'', i.e. before the explore sub-policy policies learn to enter goals. The other reward schemes do not seem successful at this challenging task without the divergence auxiliary loss.

\section{Conclusion}

Meta Reinforcement Learning has a potential for large-scale impact as a few-shot reinforcement learning algorithm could be widely applicable to many real-world domains. 
Previous work in meta reinforcement learning has almost exclusively focused on a sequential processing and conditioning of interaction episodes to produce a rapidly adaptive policy. 
The scability of this sequential approach faces several significant challenges. The most important difficulty is that as the complexity of the target tasks  increase, this might require a corresponding increase in the length and number of interactions required for optimal behaviour. As the meta-horizon grows, the gradient-based optimization procedures underlying most meta reinforcement learning algorithms can become more difficult to train. 

Motivated by this scalability issue, in this work we presented an alternative concurrent framework for meta reinforcement learning. In this ``Concurrent Meta Reinforcement Learning'' setting, interactions are processed in parallel using separate agents, with the meta-learning aspect of the framework represented by a communication protocol between the separate rollout agents. Beyond the reduction of the meta horizon to the time horizon of the environment, the multi-agent setting also afforded us the use of several specialized reward sharing schemes between the agents and the use of divergence losses to encourage a variety of agent behaviours. In contrast to previous reward schemes devised for the sequential setting~\cite{stadie2018some}, these novel reward sharing schemes can allow the agents to take high-risk actions without requiring the complete removal of external reward signal during exploration sub-episodes. We demonstrated the effectiveness of our framework on a set of challenging partially-observable maze environments, with concurrent architectures outperforming those in the sequential setting.

{\bf Acknowledgments: } This work was supported in part by the Office of Naval Research grant \#N000141812861, NSF Award IIS1763562, DARPA HR-00111890003, Apple, and Google focused award. We would also like to acknowledge NVIDIA’s GPU support.

\bibliographystyle{ACM-Reference-Format} 
\bibliography{references} 

\appendix

\section{Hyperparameter Details}
\label{sec:hyperparam}

{\bf 10-Monty-Hall:} For CMRL, we used LSTM sizes of 16 units and for baselines an LSTM with 32 units, to match parameters roughly. We did a sweep over learning rates over $\{0.00075, 0.0005, 0.00025, 0.0001\}$ for all models with 3 seeds and chose the learning rate that had highest mean performance over all seeds. For CMRL, we further did a preliminary limited sweep over divergence scale $\lambda_{div}$ and found value 0.05 with the JS divergence to work well.

{\bf 3-Color-Choice} For CMRL, we used LSTM sizes of 16 units and for baselines an LSTM with 32 units, to match parameters roughly. We did a sweep over learning rates over $\{0.0075, 0.005, 0.0025, 0.001\}$ for all models with 3 seeds and chose the learning rate that had highest mean performance over all seeds. For CMRL, we further did a preliminary limited sweep over divergence scale $\lambda_{div}$ and found value 0.0001 with the symmetric KL divergence to work well. We set $H_M = W_M = 7$ for the results.

{\bf 3-Reacher} For CMRL, we used LSTM sizes of 16 units and for baselines an LSTM with 32 units, to match parameters roughly. We did a sweep over learning rates over $\{0.0075, 0.005, 0.0025, 0.001\}$ for all models with 3 seeds and chose the learning rate that had highest mean performance over all seeds. For CMRL, we further did a preliminary limited sweep over divergence scale $\lambda_{div}$ and found value 0.001 with the JS divergence to work well.

\end{document}